\documentclass[11pt, twoside]{amsart}

\usepackage[utf8]{inputenc}
\usepackage[T1]{fontenc}
\usepackage{amssymb,amsmath,amstext}
\usepackage{hyperref, graphicx}
\usepackage{comment}
\usepackage{mathtools}
\usepackage{amsaddr}

\newtheorem{theor}{Theorem}[section]
\newtheorem{lem}[theor]{Lemma}
\newtheorem{defin}[theor]{Definition}

\newtheorem{prop}[theor]{Proposition}

\newtheorem{exam}[theor]{Example}

\newtheorem{rem}[theor]{Remark}

\newtheorem{algorithm}[theor]{Algorithm}

\numberwithin{equation}{section}

\newcommand{\mr}{\mathrm}

\newcommand{\es}{\emptyset}

\newcommand{\nts}{\negthickspace}
\newcommand{\uhrc}{\nts \upharpoonright \nts}

\newcommand{\mcA}{\mathcal{A}}
\newcommand{\mcB}{\mathcal{B}}

\newcommand{\mcE}{\mathcal{E}}
\newcommand{\mcF}{\mathcal{F}}

\newcommand{\mbV}{\mathbf{V}}
\newcommand{\mbW}{\mathbf{W}}
\newcommand{\mbX}{\mathbf{X}}
\newcommand{\mbY}{\mathbf{Y}}
\newcommand{\mbZ}{\mathbf{Z}}

\newcommand{\mbOm}{\mathbf{\Omega}}

\newcommand{\mbbG}{\mathbb{G}}
\newcommand{\mbbP}{\mathbb{P}}
\newcommand{\mbbN}{\mathbb{N}}
\newcommand{\mbbM}{\mathbb{M}}
\newcommand{\mbbR}{\mathbb{R}}

\newcommand{\rng}{\mathrm{rng}}

\title[Asymptotics for Markov logic networks]{Domain size asymptotics for Markov logic networks}

\author{Vera Koponen}

\address{Department of Mathematics, Uppsala University, Sweden.}
\email{vera.koponen@math.uu.se}

\date{25 October, 2026}

\begin{document}

\begin{abstract}
A Markov logic network (MLN) $\mbbM$ determines a probability distribution $\mbbP_n^\mbbM$
on the set $\mbW_n$ of structures,
or ``possible worlds'', with domain $\{1, \ldots, n\}$.
We study the properties of such distributions as $n$ tends to infinity.

We show that with mild assumptions on an MLN $\mbbM$ with one soft constraint with an
arbitrary positive weight the distribution $\mbbP_n^\mbbM$ will
behave quite differently from the uniform distribution $\mbbP_n^{uni}$ on $\mbW_n$ for all sufficiently large $n$.

We prove that, for every language with at least one relation symbol of arity 1, 
quantifier-free MLNs and lifted Bayesian networks (LBNs) are incomparable in the following sense:
(a) There is a quantifier-free MLN $\mbbM$ such that 
for every LBN $\mbbG$, if $n$ is sufficiently large then
$\mbbM$ and $\mbbG$ determine different distributions on $\mbW_n$.
(b) There is an LBN $\mbbG$ such that 
for every quantifier-free MLN $\mbbM$, if $n$ is sufficiently large then
$\mbbG$ and $\mbbM$ determine different distributions on $\mbW_n$.

For a language with only one relation symbol $R$ which has arity 1 and any 
quantifier-free MLN $\mbbM$
we describe of the asymptotic behaviour of
$\mbbP_n^\mbbM$ as $n \to \infty$ with respect to the most likely distribution of 
elements that satisfy $R(x)$.
If follows that the soft constraints and weights of such MLNs determine the asymptotic behaviour
of the sequence of distributions.

It is known that if $\mbbM_w$
is an MLN with one consistent soft constraint 
$\varphi(x_1, \ldots, x_k)$ with weight $w$ then, for every fixed $n$,
$\lim_{w\to\infty}$ $\mbbP_n^{\mbbM_w}\big(\forall x_1, \ldots, x_k$ $\varphi(x_1, \ldots, x_k)\big) = 1$.
We prove that if $k \geq 3$ and $\varphi(x_1, \ldots, x_k)$
expresses that $x_1$ is {\em not} connected to all of $x_2, \ldots, x_k$, then for every fixed $w \geq 0$,
$\lim_{n\to\infty}$ $\mbbP_n^{\mbbM_w}\big(\forall x_1, \ldots, x_k$ $\varphi(x_1, \ldots, x_k)\big) = 0$.
So in the limit, the weight dimension and the domain size dimension may behave completely differently.

\end{abstract}

\maketitle

\section{Introduction}

\subsection{Logic and probability}

In many situations we need to deal with objects,
properties that they may have, and relations that may exist between the objects.
Then there is also a need to express conditions, or knowledge, about these objects, properties 
and relations. This can be done in a precise and principled way by various kinds of formal
logics, e.g. first-order logic. The logical framework is also well adapted for (automated)
reasoning with certain knowledge and for deriving rules from examples, as in inductive
logic programming.

However, a purely logical framework is not able to deal with uncertainty, while 
a purely probabilistic approach to AI lacks a principled way of representing knowledge
about properties and relations of, or between, objects coming from some arbitrary (unspecified) set, 
or domain as
we will often say.
Hence there is a need of combining the framework of logic for representing certain knowledge
with concepts and methods from probability theory for dealing with uncertainty.
The field of {\em statistical relational AI (SRAI)} strives to do this
\cite{BKNP, RKNP, GT, KMG}.
In SRAI, a given domain (set), say $D$, of objects together with 
some specification of properties that these objects have and some specification of relations
between objects in $D$ is often called a {\em possible world}, which we can also think of as
a possible state of some system. 
To reuse the often used example of 
``smokers and friends'', suppose that $D = \{a, b, c\}$
where $a$, $b$ and $c$ denote some persons, $S$ denotes the property of being a smoker,
and $F$ denotes the relation of friendship between two persons
\cite{BKNP, RKNP, DL, GT, KMG, RD}.
If we specify that $a$ and $c$ are smokers, but $b$ is not (in symbols $S(a)$, $S(c)$
but $\neg S(b)$), and that $a$ and $c$ are friends (in symbols $F(a, c)$) but the other pairs are
not friends (e.g. $\neg F(a, b)$), then we get a possible world with domain $D$ for the language 
$\sigma = \{S, F\}$. In the terminology of formal logic, a possible world with domain $D$ for the
language $\sigma$ is called a {\em $\sigma$-structure}. 
In the described possible world ($\sigma$-structure) it holds that whenever
two different persons are friends then either both are smokers or none of them is, so every grounding in $D$ of the
first-order formula $\varphi(x, y) := F(x, y) \rightarrow \big(S(x) \leftrightarrow S(y)\big)$
is true in that world.
But rather than considering just one possible world, we are interested in the set,
which we denote by $\mbW_D$, of all possible worlds with domain $D$, and a probability distribution,
say $\mbbP_D$, on $\mbW_D$.
For every choice of individuals from $D$, say $a$ and $b$, we can now talk about the
probability that the grounded formula $F(a, b) \rightarrow \big(S(a) \leftrightarrow S(b)\big)$ is true
in a possible world. We can also, for example, talk about the probability that the proportion of true groundings
of $\varphi(x, y)$ in $D$ is (say) at least $2/3$.

As $\mbW_D$ is finite, 
we can in principle let $\mbbP_D$ be any possible assignment of numbers in $[0, 1]$ to all members
of $\mbW_D$ such that the sum equals one.
But then the description of $\mbbP_D$ may be as big as the cardinality of $\mbW_D$, denoted $|\mbW_D|$,
where $|\mbW_D|$ grows exponentially in $|D|$, the cardinality of $D$.
So if $D$ is large and the simplest description of $\mbbP_D$ is very large, then it will be
inefficient and difficult to reason with $\mbbP_D$.
Moreover, if $\mbbP_D$ does not reveal any regularities (patterns), then it is difficult to see if and how
knowledge about events on $\mbW_D$, using $\mbbP_D$, can be transferred (extrapolated)
to $\mbW_{D'}$ for another domain $D'$ with different cardinality and some appropriate 
distribution $\mbbP_{D'}$ on $\mbW_{D'}$.

\subsection{Knowledge transfer and lifted probabilistic graphical models}

To overcome these problems and to work with distributions that make some sense
to humans (as opposed to arbitrary distributions), researchers in SRAI have
invented precise specification languages such that each specification described by such a  language, say $\mbbG$,
tells precisely how to define a probability distribution on $\mbW_D$ 
(the set of possible worlds with domain $D$) for every domain $D$.
Such specification languages have variously been called {\em templated probabilistic graphical models},
{\em parametrized probabilistic graphical models}, or {\em lifted probabilistic graphical models},
and we will use the last expression; see e.g. \cite{BKNP, RKNP, GT, KMG}.
We will not give a precise definition of what a {\em lifted probabilistic graphical model (LPGM)} is (or should be)
but LPGMs that appear in the literature have in common that they use
some formal logical language (e.g. first-order logic), some numerical parameters,
and possibly a graph (directed or undirected) for describing (conditional) dependencies and independencies
between ``lifted/parametrized'' random variables (from the probability theoretic point of view), alternatively
relations (from the logical point of view).
{\em The essential thing in this study is that if $\mbbG$ is called an LPGM then it is only 
assumed that $\mbbG$ is a specification of 
how to define, for every finite domain $D$, a probability distribution, denoted $\mbbP_D^\mbbG$, on $\mbW_D$.}
Examples of LPGMs include Markov logic networks \cite{RD}, 
various kinds of lifted Bayesian networks \cite{CM, Jae98a, Kop20, KW2, Wei24}, and
probabilistic logic programs \cite{RS}.
More LPGMs, mostly variations of the above themes, are found in for example
\cite{RKNP, GT, KMG}.

Typically one can ``ground'' an LPGM on any finite domain $D$
and get a {\em probabilistic graphical model} on a finite set of random variables in the sense
of probability theory, as in e.g. \cite{Koller}.
Usually all groundings of a fixed LPGM will share some common regularities, or patterns,
regarding dependencies between random variables.
Example~\ref{example of MLN} and Figure~\ref{grounded MLN} illustrate this.
This makes at least some weak degree of knowledge transfer from one domain to another possible.

Let $\mbbG$ be an LPGM and $R$ a relation of arity $\nu$, say (we view properties as relations of arity 1).
Let $D$ and $D'$ be domains of different sizes and let $\bar{a} \in D^\nu$ and $\bar{a}' \in (D')^\nu$.
It is not, in general, clear if the probability $\mbbP_D^\mbbG(R(\bar{a}))$
that $\bar{a}$ satisfies $R$ (in a random possible world from $\mbW_D$)  is the same
as the probability $\mbbP_{D'}^\mbbG(R(\bar{a}'))$ that $\bar{a}'$ satisfies $R$
(in a random possible world from $\mbW_{D'}$).
The answer is even less clear if we replace $R(\bar{a})$ and $R(\bar{a}')$ by groundings $\psi(\bar{a})$
and $\psi(\bar{a}')$ of some complex formula $\psi(\bar{x})$ where the length of the sequence
$\bar{x}$ of free variables equals the length of $\bar{a}$ and $\bar{a}'$.
This problem of knowledge transfer is relevant for both inference and learning.

\subsection{Implications for learning and inference}

Let $C$ express the property of having cancer.
Suppose that $\mbbG$ is an LPGM, $D$ is a domain, $a$ a random member of $D$, and that
$\mbbP_D^\mbbG(S(a)) = 0.3$ 
(where $S(x)$ means ``$x$ smokes'') 
and $\mbbP_D^\mbbG(S(a) \wedge C(a)) = 0.1$.
This means that $\mbbG$ predicts that, on possible worlds with domain $D$, 
the probability that a smoker has cancer is $1/3$.
Let $D'$ be another domain with different size than $D$ and $a'$ a random member of $D'$.
The question is if $\mbbP_{D'}^\mbbG\big(S(a')\big) \approx 0.3$ and 
$\mbbP_{D'}^\mbbG(S(a') \wedge C(a')) \approx 0.1$.
If yes, then we can extrapolate our inference on possible worlds with domain $D$
to conclude that on possible worlds with domain $D'$ the probability that a smoker has cancer is approximately
$1/3$.
If no, then we have to compute $\mbbP_{D'}^\mbbG(S(a'))$ and $\mbbP_{D'}(S(a') \wedge C(a'))$
which, if $D'$ is large, may require large resources
(as $|\mbW_{D'}|$ is exponential in $|D'|$), and we cannot use $\mbbG$ to reason 
generally about relationships between, in this case, smoking and cancer.
Of course these issues become even more complex if we, instead of the simple first-order formulas $S(x)$
and $S(x) \wedge C(x)$, 
consider more complex formulas, possibly involving quantifiers or aggregations.

The problem of ``stability'' across domains is relevant for learning as well.
Suppose that an LPGM $\mbbG$ has been learned from examples from a domain $D$ of humans.
Then the structure and parameters/weights of $\mbbG$ have been calibrated so that $\mbbP_D^\mbbG$
ought to reflect the real distribution of events of interest in the space $\mbW_D$.
Suppose for example that, for a random $a$, $\mbbP_D^\mbbG(S(a)) = 0.3$
and $\mbbP_D^\mbbG(S(a) \wedge C(a)) = 0.1$, so $\mbbG$ predicts, on possible worlds with domain $D$,
that the probability that a smoker has cancer is $1/3$.
If $\mbbG$ is to be useful for inference in a wider context we should also 
have $\mbbP_{D'}^\mbbG(S(a')) \approx 0.3$ and $\mbbP_{D'}^\mbbG(S(a') \wedge C(a')) \approx 0.1$
for other domains $D'$ and a random member $a' \in D'$.
In fact, if $\mbbG$ is to be useful for inferences about 
probabilities of more complex events (expressed by more complex logical formulas), 
on domains of interest, these probabilities should also remain stable as we move from one domain
to another.

To better understand if probabilities defined by LPGMs are stable across domain sizes we can study the
asymptotics as the domain size tends to infinity. This will give information about how such probabilities behave
on large domains, where large domains typically require large resources (time and memory) to work with.
As the nature of the objects in the domain will not matter in this study we will often let $D := [n] := \{1, \ldots, n\}$
for some positive integer $n$. Then we let $\mbW_n := \mbW_{[n]}$ and, for an LPGM $\mbbG$,
$\mbbP_n^\mbbG := \mbbP_{[n]}^\mbbG$.
If $\varphi$ is a sentence (i.e. formula without free variables) of some formal logic (e.g. first-order logic) we can now ask if 
$\lim_{n\to\infty}\mbbP_n^\mbbG(\varphi)$ exists, where $\mbbP_n^\mbbG(\varphi)$ denotes the probability
that $\varphi$ is true in a possible world in $\mbW_n$.
If the limit exists then it follows that, for every $\varepsilon > 0$, if $n$ is large enough then 
$|\mbbP_n^\mbbG(\varphi) - \mbbP_m^\mbbG(\varphi)| < \varepsilon$ for all sufficiently large $m \geq n$.
So if $\varepsilon > 0$ is small then $\mbbP_n^\mbbG(\varphi)$ is a good approximation for $\mbbP_m^\mbbG(\varphi)$
for all $m \geq n$, which is good for extrapolation and scalability, hence also for learning and inference.
One can reason similarly for groundings of formulas $\varphi(x_1, \ldots, x_\nu)$ with free variables $x_1, \ldots, x_\nu$.

There are a number of studies giving rather general positive results regarding the
existence of limits as above for {\em directed} LPGMs, which are LPGMs which use a directed acyclic graph
to describe (in)dependencies between relations, including 
\cite{CM, Jae98a, Kop20, Kop24, KT, KW1, KW2, Wei21, Wei24}.
But for {\em undirected} LPGMs such as Markov logic networks (MLNs) not much is known.
However, there are studies showing that some MLNs behave in an undesirable way with respect to knowledge transfer.
In particular there are examples of MLNs $\mbbM$ such that the numerical parameters of $\mbbM$
have negligible influence on the probability of some relevant events when the domain size is large
\cite{Mittal, Poole, Wei25}. 
Therefore such numerical parameters which have been learned from examples on one domain to 
reflect probabilities on that domain lose their meaning on sufficiently large domains.

Although directed LPGMs seem to be better adapted for knowledge transfer, undirected LPGMs have
the advantage of being able to describe ``cyclic'' probabilistic dependencies, in particular 
situations where the probability that $P(a)$ holds depends on whether $P(b)$ holds for $b \neq a$.
For example, if $P(x)$ means ``$x$ has the flu'', then whether or not $a$ has the flu may influence the probability that $b$ has the flu.
If less than 1 percent of the population has a flu then the probability that a random person has a flu may be quite low, 
but if more than 30 percent of the population has a flu then the probability that a random person has it ought to be higher.
Such a dependency of the event that $a$ has a flu on the event that some subset of other persons have a flu can be 
described by an MLN over the language $\{P\}$, but not by for example lifted Bayesian networks 
(which are directed LPGMs) over $\{P\}$ in the sense of
Definition~\ref{definition of LBN} below, as we will see.
We will also see that MLNs are not in general more expressive than lifted Bayesian networks.

\subsection{Markov logic networks}

The concept of a Markov logic network (MLN), conceived by Richardson and Domingos \cite{RD} in 2006,
is an attractive LPGM since it has a simple intuitive definition. The definition of how it determines a probability distribution
on $\mbW_D$ for any finite domain $D$ is straightforward as well.
Algorithms for learning MLNs and making inferences with MLNs are discussed in e.g.
\cite{BKNP, RKNP, DL, KMG, RD}.
But there is no known algorithm such that there is a polynomial $f(x, y, z)$ such that for
every Markov logic network $\mbbM$, every first-order sentence $\varphi$ and every finite domain $D$,
the algorithm computes (or estimates with as high accuracy as we like) $\mbbP_D^\mbbM(\varphi)$
in time bounded by $f(|D|, |\mbbM|, |\varphi|)$, where $|\mbbM|$ and $|\varphi|$ denote the
length of $\mbbM$ and $\varphi$, respectively, viewed as sequences of symbols.

A {\em Markov logic network} (MLN) over a language $\sigma$
is a finite set of pairs $(\varphi(x_1, \ldots, x_\nu), w)$ 
where $\varphi(x_1, \ldots, x_\nu)$ is a first-order formula
that uses only relation symbols from $\sigma$ and possibly the identity symbol `=', and $w$ is a 
nonnegative real number, 
called the {\em weight} of the formula.
The formula $\varphi(x_1, \ldots, x_\nu)$ in such a pair is called a {\em soft constraint}
and the intuition is that its weight $w$ gives a measure of how likely it is that this soft constraint
is satisfied by a $\nu$-tuple of elements from a domain. 
Somewhat more precisely, if $w > 0$ and 
$\mcA$ and $\mcB$ are two possible worlds (i.e. $\sigma$-structures) with the same domain
$D$ say, then, other things equal, if the number of $\bar{d} \in D^\nu$ that satisfy
$\varphi(x_1, \ldots, x_\nu)$ in $\mcA$ is larger than the number of $\bar{d} \in D^\nu$ 
that satisfy 
$\varphi(x_1, \ldots, x_\nu)$ in $\mcB$, then $\mcA$ is more likely than $\mcB$ according to 
the distribution on $\mbW_D$ determined by $\mbbM$ (according to
Definition~\ref{definition of distribution defined by an MLN}).
Moreover, the larger $w$ is the more are violations of 
$\varphi(x_1, \ldots, x_\nu)$ penalized in the sense of making the possible world less likely.

\begin{exam}\label{example of MLN}{\rm
Let $\sigma = \{S, C, F\}$, with the same meanings of $S$, $C$ and $F$ as before, and
\[
\mbbM := \big\{\big(F(x, y) \rightarrow (S(x) \leftrightarrow S(y)), 2\big), \ 
\big(S(x) \to C(x), 3\big)\big\}.
\]
The first pair of $\mbbM$ stipulates that there is a tendency that if two persons are friends then 
either both smoke or none of them smoke. The second pair of $\mbbM$ stipulates that 
people who smoke have a higher probability of having cancer.  
If we would change, say, the second weight from 3 to 4, then the new MLN would
make the correlation between smoking and cancer stronger.
}\end{exam}

\begin{figure}[h]

\caption{\label{grounded MLN}
\small Let $\mbbM$ be the MLN from Example~\ref{example of MLN} and let $D = \{a, b, c\}$.
Then the graph structure of the Markov random field on $D$ that is obtained from $\mbbM$ is
described below, except for the vertices $F(a, a)$, $F(b, b)$ and $F(c, c)$
which are omitted. Each atomic formula of the Markov random field, e.g. $S(a)$, $F(a, b)$ etc,
can be viewed as a 0/1-valued random variable, and edges indicate dependencies which may
go in both directions. Cliques which correspond to a grounded soft contraint have a potential associated to them.
For example, the clique $\{S(a), S(b), F(a, b)\}$ has potential $2^2$.
Note that the clique $\{S(a), S(b), S(c)\}$ does not correspond to a soft constraint, because there is
no grounding $\chi$ of a soft constraint of $\mbbM$ such that all $S(a), S(b), S(c)$ are subformulas 
of $\chi$. By default we give this clique the potential $2^0 = 1$, intuitively meaning that it
has no influence. }
\label{grounded_MLN}
\begin{center}
\includegraphics[scale=1.1]{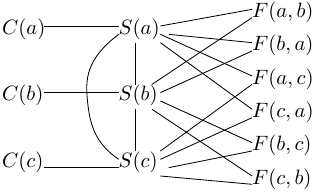}
\end{center}

\end{figure}

\noindent
An MLN $\mbbM$ with quantifier-free soft constraints
can be seen as a template for constructing a {\em Markov random field}
(also called {\em Markov network})
on any finite set $D$.
The procedure is as follows, and is illustrated by Figure~\ref{grounded_MLN}:
The set $V$ of vertices of the Markov random field on $D$ is the set of all groundings in $D$ of atomic subformulas
of the soft constraints of $\mbbM$. Each member $\psi \in  V$ 
can be seen as a 0/1-valued random variable indicating
whether $\psi$ is true or not.
If $\psi, \theta \in V$ then we draw an undirected edge between $\psi$ and $\theta$ {\em if}
there is some grounding, say $\chi$, of a soft constraint of $\mbbM$ such that both $\psi$ and $\theta$
are subformulas of $\chi$. 
Thus, for every grounding $\chi$ of a soft constraint the set of all atomic subformulas of $\chi$
will form a clique (complete graph) and this clique is given a so called {\em potential} which is
equal to 2 (or some other base > 1) raised to the weight of the soft constraint that $\chi$ grounds.

\subsection{Contributions}

As the only thing about a domain that will matter in this study is its size (cardinality) we will only consider
domains of the form $D := [n] := \{1, \ldots, n\}$ for positive integers $n$.
By $\mbW_n$ we denote the set of all $\sigma$-structures, for the language $\sigma$ of consideration,
with domain $[n]$.
If $\mbbM$ is an MLN then $\mbbP_n^\mbbM$ denotes the probability distribution on $\mbW_n$
that is determined by $\mbbM$ according to 
Definition~\ref{definition of distribution defined by an MLN}.

Let $\mbbM = \{(\varphi(x_1, \ldots, x_\nu), w)\}$ be an MLN.
From Definition~\ref{definition of distribution defined by an MLN}
it follows that if $w = 0$ then, for all $n \in \mbbN^+$, $\mbbP_n^\mbbM = \mbbP_n^{uni}$ 
where $\mbbP_n^{uni}$ denotes the uniform probability distribution on $\mbW_n$.
One may therefore think that if $w > 0$ is only very slightly larger than 0, 
then $\mbbP_n^\mbbM$ does not differ that much from $\mbbP_n^{uni}$.
However, from an asymptotic perspective the contrary is typically true, as stated by
Theorem~\ref{with high probability there are few violations}:
if $w > 0$ and $\varphi(x_1, \ldots, x_\nu)$ satisfies some mild conditions,
then for all $n$ there is $\mbX_n \subseteq \mbW_n$ such that 
$\lim_{n\to\infty} \mbbP_n^\mbbM(\mbX_n) = 1$ and $\lim_{n\to\infty} \mbbP_n^{uni}(\mbX_n) = 0$.
Theorem~\ref{with high probability there are few violations} 
also shows that $\mbbP_n^\mbbM$ and $\mbbP_n^{uni}$ behave in opposite ways (for large $n$)
with respect to the number
of $\nu$-tuples that satisfy/falsify $\varphi(x_1,\ldots, x_\nu)$ in a random member of $\mbW_n$.

The next set of results,
in Section~\ref{Colour distributions expressible with quantifier-free MLNs}, 
concern the question whether MLNs can define more or less
sequences $(\mbbP_n : n \in \mbbN^+)$ of distributions on $\mbW_n$, $n \in \mbbN^+$,
than {\em lifted Bayesian networks (LBNs)}
(Definition~\ref{definition of LBN}) can do, or if the two formalisms are ``incomparable'' in the sense 
that none of them is more expressive than the other.
Theorem~\ref{MLN versus LBN generally}
shows that if the language $\sigma$ contains at least one relation symbol of arity 1,
then MLNs and LBNs over $\sigma$ are incomparable.
An analogous incomparability result is known since long time ago for propositional Markov networks and
propositional Bayesian networks (see e.g. \cite[Chapter~3]{Pea}), but to my best knowledge this study
provides the first extension to the context of first-order logic.
Also, a detailed study of quantifier-free MLNs $\mbbM$ over $\sigma := \{R\}$ where $R$ has arity 1
is conducted. It characterizes asymptotic behaviour of 
the sequence $(\mbbP_n^\mbbM : n \in \mbbN^+)$.
It shows that the behaviour of $\mbbP_n^\mbbM$ stabilizes as $n \to \infty$, is predictable, 
can be inferred by small computational resources,
and the weights of the soft constraints of $\mbbM$
determine the behaviour (see Theorems~\ref{MLN with two convergence points},
\ref{characterization of quantifier-free MLNs for colourings}, and~\ref{limits of MLNs}).

It is known that for an MLN $\mbbM_w = \{(\varphi(x_1, \ldots, x_k), w)\}$, if
$\varphi(x_1, \ldots, x_k)$ is satisfiable in some member of $\mbW_m$ then
$\lim_{w \to \infty} \mbbP_m^{\mbbM_w}(\forall x_1, \ldots, x_k \varphi(x_1, \ldots, x_k)) = 1$.
In Section~\ref{graphs with maximum degree}
it is demonstrated that if $k \geq 3$, $\varphi(x_1, \ldots, x_k)$
expresses that $x_1$ is {\em not} a neighbour to all $x_2, \ldots, x_k$ and 
$\mbbM_w = \{(\varphi(x_1, \ldots, x_k), w)\}$, then, for any fixed $w \geq 0$,
$\lim_{n \to \infty} \mbbP_n^{\mbbM_w}(\forall x_1, \ldots, x_k \varphi(x_1, \ldots, x_k)) = 0$.
Thus the limit as $w \to \infty$ and the limit as $n \to \infty$ may be as different as possible,
so the weight and domain size dimensions may be completely detached.

\subsection{Related work and discussion}\label{Related work}

There is a fairly large number of rather general (and mostly recent) results concerning 
convergence for broad classes of logical formulas with respect to distributions generated by different kinds of
{\em directed acyclic LPGMs} such as
relational Bayesian networks \cite{Jae98a},
relational Bayesian network specifications \cite{CM},
lifted Bayesian networks \cite{Kop20, KW1},
probabilistic logic programs \cite{Wei21},
PLA-networks \cite{Kop24, KT, KW2}, and
functional lifted Bayesian networks \cite{Wei24}.
There are also convergence results related to graph neural networks
\cite{Adam-Day1, Adam-Day2, Adam-Day3}
and multilayer feedforward neural networks \cite{KopNeu}.

Regarding convergence results
for distributions generated by {\em undirected} LPGMs I am only aware of the articles \cite{JBB, Mittal, Poole, Wei25}.
Poole, Buchman, Kazemi, Kersting, and Natarajan \cite{Poole}
followed up by Mittal, Bhardwaj, Gogate, and Singla \cite{Mittal} and
by Weitkämper \cite{Wei25} showed the following:
If $\mbbM$ is an MLN with a single soft constraint
and $\varphi$ denotes a specific grounded atomic formula, then $\lim_{n\to\infty} \mbbP_n^\mbbM\big(\varphi\big)$ 
need {\em not} depend on the weight of the soft constraint of $\mbbM$.
In these studies the soft constraint considered seems a bit contrived as it does not seem to have any intuitive meaning.
In any case, since the limit (as above) does not depend on the weight of the soft constraint, the weight loses its
influence with respect to $\mbbP_n^\mbbM\big(\varphi\big)$ when $n$ is large enough.

To remedy the loss of influence of the weight of the soft constraint, for large $n$, a couple of modifications of
Markov logic networks have been proposed
by  Jain, Barthels, and Beetz (Adaptive MLNs) \cite{JBB},
and by Mittal, Bhardwaj, Gogate, and Singla (domain-size aware MLNs) \cite{Mittal}.
The common idea of these modifications is the following:
Suppose that $\mbbM$ is an (ordinary) MLN with weights learned on a domain with size $m$.
For simplicity suppose that $\mbbM$ has only one soft constraint with learned weight $w$.
Now we construct a {\em rescaled MLN}  $\mbbM'$ 
which replaces the constant weight $w$ 
by a function $f_w$, that depends on $w$, from the positive integers to the nonnegative reals.
For each positive integer $n$, $\mbbM'$ determines a probability distribution on $\mbW_n$ by taking $\mbbM$,
replacing its weight $w$ by $f_w(n)$ and using the so modified $\mbbM$ to determine a distribution on $\mbW_n$
in the usual way for MLNs.
Thus, the weight of $\mbbM'$ is not a constant but a function of $n$ that depends on $w$.
In \cite{JBB} and \cite{Mittal}
it is shown that, for the earlier considered soft constraints in \cite{Poole}, and ground formulas $\varphi$ (matching
the respective soft constraint)
and with the respective rescaling $\mbbM'$ of $\mbbM$,
the limit probability of $\varphi$, using $\mbbM'$, exists and {\em depends} on $w$, the weight of the single
soft constraint of the original MLN $\mbbM$.

However, Weitkämper \cite{Wei25} later showed that if we consider {\em another} ground atomic formula, say $\psi$, 
from the same example of MLN,
then the limit probability of $\psi$ (using the rescaled $\mbbM'$) exists and does {\em not} depend on the weight $w$
of the original $\mbbM$. So the approach of rescaled MLNs in \cite{JBB, Mittal} 
does not, in general, solve the problem of 
the weights
becoming irrelevant in the limit.
To complicate the picture further, it is clear from the results in 
Section~\ref{Colour distributions expressible with quantifier-free MLNs}
that the weights of the MLNs considered there {\em do influence}  the limit probabilities of ground atomic formulas.

Besides that there seems to be no known rescaling that always gives dependence on the weights in 
the limit of ground atomic formulas, let alone for logical formulas in general, 
and besides that sometimes the weights matter without rescaling
(as demonstrated in Section~\ref{Colour distributions expressible with quantifier-free MLNs}), 
the question is also how to motivate that any kind of rescaled weights  are the ``right ones''.
Or why should the limit obtained with some of the suggested rescalings in each of a few examples be considered the 
``right'' limit, especially if it is not made clear in 
a sufficiently general setting how
the limit of the rescaled MLN depends on the original MLN? 
A criterion for a ``good rescaling'' could arguably be that if (for simplicity) $\mbbM$ is an 
MLN with only one soft constraint with weight $w$
which has been learned from a domain of size $m$, $\mbbP_n^\mbbM$ is the distribution determined by $\mbbM$ on $\mbW_n$,
$\mbbM'$ is the rescaled MLN (obtained from $\mbbM$),
$\mbbP_n^{\mbbM'}$ is the distribution on $\mbW_n$ determined by $\mbbM'$,
and $\varphi$ is a ground atomic formula,
then $\mbbP_m^\mbbM\big(\varphi\big)$ and 
$\mbbP_n^{\mbbM'}\big(\varphi\big)$
should be approximately the same for all sufficiently large $n$; the closer the better.

But given the uncertainty about what would be the ``right''
rescaling of an ordinary MLN to get asymptotic probabilities where the original 
(learned and constant)
weights matter in a sensible way, it seems of interest to further study  the asymptotic properties of distributions $\mbbP_n^\mbbM$
determined by ordinary MLNs  $\mbbM$, with various types of soft constraints,
 to get a better idea of the asymptotic behaviour of MLNs with respect to the domain size.
Are there patterns and
to which extent would such patterns guide our use of MLNs, or of their rescalings?

\subsection{Notation and terminology}

\noindent
$\mbbN$ and $\mbbN^+$ denote the set of nonnegative, respectively, positive integers.
Let $\exp_2(x) := 2^x$.
If $S$ is a set then $|S|$ denotes its cardinality, also informally called its size.
If $T$ is also a set then $S \setminus T$ is the set of elements that belong to $S$ but not to $T$.
Finite sequences will be denoted by $\bar{s}$ for some letter $s$, the length of $\bar{s}$ will be denoted by $|\bar{s}|$,
and the set of elements occurring in $\bar{s}$ will be denoted by $\rng(\bar{s})$.

Basic knowledge about first-order logic will be assumed (see e.g. \cite{EF, Gor}).
If a sequence of logical variables is denoted by $\bar{x}$ (say) then it is assumed that all entries in $\bar{x}$ are different
(i.e. if $\bar{x} = (x_1, \ldots, x_k)$ then $x_i \neq x_j$ whenever $i \neq j$).
By a {\em language} (also called {\em signature} or {\em vocabulary}) we mean a finite set of relation symbols $\sigma$ 
where all relation symbols in $\sigma$ have arity $\geq 1$
and the identity symbol `$=$' does not belong to $\sigma$.
If $R \in \sigma$ then $\mr{ar}(R)$ denotes its arity.
If $\sigma$ is a language then $FO(\sigma)$ is the set of first-order formulas that can be formed by using symbols in $\sigma$
and the identity symbol `$=$'.
If a formula in $FO(\sigma)$ is denoted by $\varphi(\bar{x})$ (where $\bar{x}$ is a finite sequence of logical variables)
then it is assumed that every free variable in the formula denoted by $\varphi(\bar{x})$ occurs in $\bar{x}$ and that
every variable that occurs in $\bar{x}$ is a free variable of the formula denoted by $\varphi(\bar{x})$.
A first-order formula is called {\em quantifier-free} if it contains no quantifier, that is, if it is composed by using only
atomic formulas and connectives among $\neg, \wedge, \vee, \rightarrow$, and $\leftrightarrow$.
First-order structures will be denoted by calligraphic letters $\mcA$, $\mcB$, etc. and unless something else
is said their domains are denoted by the corresponding noncalligraphic letters $A$, $B$, etc.
If $\varphi(x_1, \ldots, x_k) \in FO(\sigma)$ and $\mcA$ is a $\sigma$-structure with domain $A$ (i.e. an interpretation of all 
relations in $\sigma$ on the domain $A$), then we let the notation `$\varphi(\mcA)$' denote 
the set of $k$-tuples of elements from the domain of $\mcA$ that satisfy $\varphi(x_1, \ldots, x_k)$, or equivalently,
\[
\varphi(\mcA) := \{ (a_1, \ldots, a_k) \in A^k : \mcA \models \varphi(a_1, \ldots, a_k)\}.
\]
If $\varphi$ is a sentence (formula without free variables) and $\mcA \models \varphi$,
then we let $\varphi(\mcA)$ be the set containing only the empty sequence, and otherwise we let $\varphi(\mcA)$ be the empty set.
If $R \in \sigma$ is a relation symbol then $R^\mcA$ denotes its interpretation in the $\sigma$-structure $\mcA$,
and it follows that $R(\mcA) = R^\mcA$.
Let $R$ be a $k$-ary relation on a set $A$. We say that $R$ is {\em irreflexive and symmetric}
if the following holds:
(a) $R(a_1, \ldots, a_k)$ implies that all $a_1, \ldots, a_k$ are different, and (b)
if $R(a_1, \ldots, a_k)$ holds then, for every permutation $\tau$ of $\{1, \ldots, k\}$,
$R(a_{\tau(1)}, \ldots, a_{\tau(k)})$ holds.

Let $\mcA$ be a $\sigma$-structure and suppose that $\tau \subset \sigma$.
The {\em reduct} of $\mcA$ to $\tau$ is the (unique) $\tau$-structure $\mcB$ such that
$A = B$ ($\mcA$ and $\mcB$ have the same domain) and for all $R \in \tau$,
$R^\mcB = R^\mcA$ ($R$ is interpreted in the same way in $\mcA$ and $\mcB$).
The reduct of $\mcA$ to $\tau$ may be denoted by $\mcA \uhrc \tau$.
If $\mcB$ is the reduct of $\mcA$ to $\tau$ then we also call $\mcA$ an {\em expansion} of $\mcB$ to $\sigma$.
A $\sigma$-structure $\mcB$ is called a {\em substructure} of a $\sigma$-structure $\mcA$ if
$B \subseteq A$ and for all $R \in \sigma$, $R^\mcB =  R^\mcA \cap B^{\mr{ar}(R)}$.
An {\em isomorphism} from a $\sigma$-structure $\mcA$ to a $\sigma$-structure $\mcB$ is
a bijective function $f : A \to B$ such that for all $R \in \sigma$ and all 
$a_1, \ldots, a_{\mr{ar}(R)} \in A$,
$\mcA \models R(a_1, \ldots, a_{\mr{ar}(R)})$ if and only if $\mcB \models R(f(a_1), \ldots, f(a_{\mr{ar}(R)}))$.
(In this case the inverse of $f$ is an isomorphism from $\mcB$ to $\mcA$.)
We call $\mcA$ and $\mcB$ {\em isomorphic} if there is an isomorphism from one structure to the other.

For all $n \in \mbbN^+$ we let $[n] := \{1, \ldots, n\}$.
If $\sigma$ is a language then $\mbW_n$ denotes the set of all $\sigma$-structures with domain $[n]$.
We assume that for each $R \in \sigma$ it is specified if $R$ can be interpreted as any relation (of arity matching that of $R$) or if
$R$ must always be interpreted as an irreflexive and symmetric relation.

\section{Markov logic networks versus the uniform distribution}

\noindent
Let $\sigma$ be a nonempty language.
In this section we define the notion of a Markov logic network over $\sigma$ and the probability 
distribution that it determines on $\mbW_n$ for each $n \in \mbbN^+$.
It will be evident that if $\mbbM = \{(\varphi(x_1, \ldots, x_\nu), w)\}$ is an MLN and $w = 0$,
then the distribution $\mbbP_n^\mbbM$ on $\mbW_n$ determined by $\mbbM$
is the uniform probability distribution.
In order to understand the flexibility of MLNs, one can now ask the question if it is possible to choose 
$w > 0$ so that, asymptotically, $\mbbP_n^\mbbM$ differs only ``moderately'' from the uniform probability distribution on $\mbW_n$ which we denote by $\mbbP_n^{uni}$.
A more precise version of the question is: Is it possible to choose $w > 0$ so that 
there is $0 < d < 1$ such that for all sufficiently large $n$, 
if $\mbX \subseteq \mbW_n$ then $\big|\mbbP_n^\mbbM(\mbX) - \mbbP_n^{uni}(\mbX)\big| \leq d$.
Theorem~\ref{with high probability there are few violations}
shows that if $\varphi(x_1, \ldots, x_\nu)$ satisfies 
some mild assumptions then this is {\em not} possible, in fact (as stated by the theorem),
for all $n \in \mbbN^+$, there is $\mbX_n \subseteq \mbW_n$
such that 
\[
\lim_{n\to\infty} \mbbP_n^\mbbM(\mbX_n) = 1 \ \ 
\text{ and } \ \ \lim_{n\to\infty} \mbbP_n^{uni}(\mbX_n) = 0.
\]
\noindent
Theorem~\ref{with high probability there are few violations}
gives more information about how $\mbX_n$ can be chosen.
It shows that $\mbbP_n^\mbbM$ and $\mbbP_n^{uni}$ behave in opposite ways
with respect to the probability of sampling a possible world with very few $\nu$-tuples
violating the soft constraint.

Finally in this section, 
we give a technical (but useful) lemma that states that for every quantifier-free MLN $\mbbM$ over $\sigma$
there is a quantifier-free MLN $\mbbM'$ such that, for all $n$, $\mbbM$ and $\mbbM'$ determine the same
distribution on $\mbW_n$ and $\mbbM'$ has a specific ``normal form''.
This lemma will be used in Section~\ref{Colour distributions expressible with quantifier-free MLNs}

\begin{defin}\label{definition of MLN}{\rm
A {\em Markov logic network (MLN) over $\sigma$} is a finite set $\mbbM$ of pairs of the form $(\varphi(\bar{x}), w)$
where $\varphi(\bar{x}) \in FO(\sigma)$, $\bar{x}$ is a sequence of distinct variables, and $w$ is a non-negative real number.
For each $(\varphi(\bar{x}), w) \in \mbbM$, we call $\varphi(\bar{x})$ a {\em soft constraint} (of $\mbbM$) 
and $w$ its {\em weight}; we also call $|\bar{x}|$ the {\em arity} of the soft constraint $\varphi(\bar{x})$.
If for every $(\varphi(\bar{x}), w) \in \mbbM$ the formula $\varphi(\bar{x})$ is quantifier-free we call $\mbbM$ a
{\em quantifier-free} MLN.
}\end{defin}

\noindent
Note that if $\mbbM$ is an MLN and $(\varphi(\bar{x}), w) \in \mbbM$ and $(\psi(\bar{y}), v) \in \mbbM$ are different pairs,
then $\bar{x}$ and $\bar{y}$ may be different sequences of variables and they may have different lengths.
We allow $\bar{x}$ in $\varphi(\bar{x})$ to be empty in which case $\varphi$ is a sentence (a formula without free variables).
Recall that if $\varphi(\bar{x})$ is a first-order formula and $\mcA$ is a finite structure,
then $|\varphi(\mcA)|$ is the number of ordered $|\bar{x}|$-tuples of elements from the domain of $\mcA$ that satisfy $\varphi(\bar{x})$
in $\mcA$.

\begin{defin}\label{definition of distribution defined by an MLN}{\rm
Let $\mbbM = \{(\varphi_1(\bar{x}_1), w_1), \ldots, (\varphi_t(\bar{x}_t), w_t) \}$ be an MLN and let $n \in \mbbN^+$.
\begin{align*}
&\text{For every $\mcA \in \mbW_n$, define } \ \mu_n^\mbbM(\mcA) = 
\exp_2\Big(\sum_{i=1}^t w_i |\varphi_i(\mcA)| \Big), \\
&\text{and for every $\mbX \subseteq \mbW_n$, define } \ \mu_n^\mbbM(\mbX) = 
\sum_{\mcA \in \mbX} \mu_n^\mbbM(\mcA), \ \text{ and} \\
&\mbbP_n^\mbbM\big(\mbX\big) = \frac{\mu_n^\mbbM(\mbX)}{\mu_n^\mbbM(\mbW_n)}.
\end{align*}
Then $\mbbP_n^\mbbM$ is a probability distribution on $\mbW_n$ which we call the 
{\em distribution (on $\mbW_n$) determined by $\mbbM$}.
}\end{defin}

\begin{rem}\label{remark on MLNs}{\rm
(a) If $\mbbM$ is the empty set, then the ``empty'' sum has value zero and it follows that $\mbbP_n^\mbbM$ is the
uniform probability distribution on $\mbW_n$.\\
(b) If all weights are zero, then $\mbbP_n^\mbbM$ is the uniform probability distribution. \\
(c) Recall that from the notational conventions it follows that, if some $\varphi_i$ 
in Definition~\ref{definition of distribution defined by an MLN} is a sentence, that is, if $\bar{x}_i$ is empty,
then $|\varphi_i(\mcA)| = 1$ if $\mcA \models \varphi_i$ and $|\varphi_i(\mcA)| = 0$ otherwise.\\
(d) The choice of 2 as a base of exponentiation is irrelevant for the results to be presented (as long as the base is larger than 1),
but will make some computations simpler.
}\end{rem}

\begin{defin}\label{definition of asymptotically persistent}{\rm
Let $\varphi(\bar{x}) \in FO(\sigma)$ where $\bar{x} = (x_1, \ldots, x_l)$ is a sequence
of distinct variables.\\
(1) We call $\varphi(\bar{x})$ {\em asymptotically persistent} if 
\[
\lim_{n\to\infty} 
\frac{\big|\big\{ \mcA \in \mbW_n : \mcA \models \exists \bar{x} \varphi(\bar{x}) \big\}\big|}
{|\mbW_n|} \neq 0.
\]
(2) For all $n \in \mbbN^+$, $\alpha \geq 0$ and $\varepsilon > 0$, define
\[
\mbX_n^{\alpha, \varepsilon}(\varphi) :=
\big\{ \mcA \in \mbW_n : (\alpha - \varepsilon)n^l \leq |\varphi(\mcA)| \leq (\alpha + \varepsilon)n^l \big\}.
\]
(3) Let $\mr{diff}(x_1, \ldots, x_l)$ denote the formula $\bigwedge_{1 \leq i < j \leq l} x_i \neq x_j$.
}\end{defin}

\noindent
If $\varphi(\bar{x})$ is quantifier-free and satisfiable then it follows from e.g. \cite{KL}
that $\varphi(\bar{x})$ is asymptotically persistent.
In general, if $\varphi(\bar{x})$ is {\em not} asymptotically persistent then it is satisfiable
in only a vanishing proportion of structures in $\mbW_n$ as $n\to\infty$, so 
as a soft constraint $\varphi(\bar{x})$ is quite restrictive.
In order to prove the second part of Theorem~\ref{with high probability there are few violations} we
need the following result which is a combination of results from \cite{Gle} and \cite{KL}.

\begin{theor}\label{typical proportion of tuples satisfying a formula} 
{\rm (Combination of results \cite{Gle} and \cite{KL})} 
Let $\bar{x} = (x_1, \ldots, x_l)$ where $l \geq 1$, let $\varphi(\bar{x}) \in FO(\sigma)$ and suppose that
$\varphi(\bar{x}) \wedge \mr{diff}(\bar{x})$ is asymptotically persistent.
Then there is $0 < \alpha \leq 1$ such that for every $\varepsilon > 0$ there is $c > 0$ such that for all sufficiently large $n$,
\begin{equation}\label{convergence of phi}
\frac{\big| \mbX_n^{\alpha, \varepsilon}(\varphi) \big|}
{\big|\mbW_n\big|}
\ \geq \ 1 - e^{-cn}.
\end{equation}
If also $\neg\varphi(\bar{x}) \wedge \mr{diff}(\bar{x})$ is asymptotically persistent, then $\alpha < 1$.
\end{theor}

\noindent
{\bf Proof.}
The results of \cite{Gle} and \cite{KL} hold no matter if some relation symbols are always interpreted as
irreflexive and symmetric relations, or not, so we need not worry about this.
Let $\varphi(\bar{x}) \in FO(\sigma)$ be as assumed.
By \cite{Gle} there is a quantifier-free $\psi(\bar{x}) \in FO(\sigma)$ such that
if 
\[
\mbZ_n = \{\mcA \in \mbW_n : 
\mcA \models \forall \bar{x}\big(\big[\varphi(\bar{x}) \wedge \mr{diff}(\bar{x})\big] \leftrightarrow \psi(\bar{x})\big)\}
\]
then there is $d > 0$ such that 
$\frac{|\mbZ_n|}{|\mbW_n|} \geq 1 - e^{-dn}$ for all sufficiently large $n$.
Every quantifier-free formula is logically equivalent to a disjunctive normal form 
(see e.g. \cite{Gor}).
Therefore we may assume that $\psi(\bar{x})$ is a disjunctive normal form $\bigvee_{k=1}^s \theta_k(\bar{x})$
where each $\theta_k(\bar{x})$ is a consistent conjunction of $\sigma$-literals.
By adding more such conjunctions to the disjunction, if necessary, we may assume that each $\theta_k(\bar{x})$
is a maximal consistent conjunction of $\sigma$-literals using the variables in $\bar{x}$.
It follows that for all $k$, $\theta_k(\bar{x}) \models \mr{diff}(\bar{x})$.

In general we may have the problem that the disjunction $\bigvee_{k=1}^s \theta_k(\bar{x})$
is empty, in which
case the disjunction is interpreted as a formula which is always false, and in this case
$\mcA \models \neg \exists \bar{x}\big(\varphi(\bar{x}) \wedge \mr{diff}(\bar{x})\big)$ for all $\mcA \in \mbZ_n$.
However, the assumption that $\varphi(\bar{x})  \wedge \mr{diff}(\bar{x})$
is asymptotically persistent implies that there is $\beta > 0$ such that if 
\[
\mbV_n := \big\{ \mcA \in \mbW_n :
 \mcA \models \exists \bar{x}\big(\varphi(\bar{x})  \wedge \mr{diff}(\bar{x})\big) \big\}
\]
then there are infinitely many $n$ such that $\frac{|\mbV_n|}{|\mbW_n|} > \beta$.
As $\lim_{n\to\infty}\frac{|\mbZ_n|}{|\mbW_n|} = 1$ it follows that there are infinitely many $n$
such that $\mbV_n \cap \mbZ_n \neq \es$.
For any such $n$ take $\mcA \in \mbV_n \cap \mbZ_n$.
As $\mcA \in \mbV_n$ it follows that there is $\bar{a} \in [n]^l$  such that 
$\mcA \models \varphi(\bar{a}) \wedge \mr{diff}(\bar{a})$.
As $\mcA \in \mbZ_n$ we have 
\[
\mcA \models \forall \bar{x}\bigg(\big[\varphi(\bar{x}) \wedge \mr{diff}(\bar{x})\big] \leftrightarrow 
\bigvee_{k=1}^s \theta_k(\bar{x})\bigg).
\]
Hence we get $\mcA \models \theta_k(\bar{a})$ for some $k$, so the disjunction is not empty.

It follows from results 
in \cite{KL} that, for every $k = 1, \ldots, s$, there is $0 < \alpha_k \leq 1$ such that for all $\varepsilon > 0$
there is $c_\varepsilon > 0$ (depending only on $\varepsilon$) such that for all sufficiently large $n$,
\begin{equation}\label{convergence of the thetas}
\frac{\big| \mbX_n^{\alpha_k, \varepsilon}(\theta_k) \big|}{|\mbW_n|}
 \ \geq \ 1 - e^{-c_\varepsilon n}.
\end{equation}
(To be more precise, the above is a consequence of Lemma~3.2 and Lemmas~4.5 and 4.6 in \cite{KL},
including their proofs to get the exponentially fast convergence to 1.)
With $\alpha = \alpha_1 + \ldots + \alpha_s$ we get~(\ref{convergence of phi}) (for some $c > 0$).

Now suppose that $\neg\varphi(\bar{x}) \wedge \mr{diff}(\bar{x})$ is asymptotically persistent.
Then there is a maximal consistent conjunction of $\sigma$-literals using the variables in $\bar{x}$, say $\theta(\bar{x})$
such that $\theta(\bar{x}) \models \mr{diff}(\bar{x})$ and
$\theta(\bar{x}) \wedge \theta_k(\bar{x})$ is inconsistent for all $k = 1, \ldots, s$.
Then~(\ref{convergence of the thetas}) holds with $\theta(\bar{x})$ in place of $\theta_k(\bar{x})$ and
some $0 < \gamma \leq 1$ in place of $\alpha_k$. Consequently $\alpha = \alpha_1 + \ldots + \alpha_s < 1$.
\hfill $\square$

\begin{defin}\label{definition of Y...}{\rm
If $\varphi(\bar{x}) \in FO(\sigma)$ then, for all $n \in \mbbN^+$ and reals $m \geq 0$, we define
\[
\mbY_n^m(\varphi) := \big\{\mcA \in \mbW_n : |\neg\varphi(\mcA)| \leq m \big\}.
\]
}\end{defin}

\noindent
Hence $\mbY_n^m(\varphi)$ consists of all $\mcA \in \mbW_n$ such that at most $m$ $|\bar{x}|$-tuples of elements from
$[n]$ ``violate'' $\varphi(\bar{x})$ in $\mcA$.
In particular, $\mbY_n^0(\varphi)$ consists of all $\mcA \in \mbW_n$ such that 
no tuple violates $\varphi(\bar{x})$ in $\mcA$,
or equivalently, $\mcA \models \forall \bar{x} \varphi(\bar{x})$.

\begin{rem}\label{remark on hereditary classes}{\rm 
In the next theorem we will assume that 
\begin{equation}\label{the soft constraint is not too strong}
\big|\mbY_n^0(\varphi)\big| \geq 2^{cn} \ \text{ for some fixed $c > 0$ and all 
sufficiently large $n$.}
\end{equation}
This is not a very restrictive condition, but a possibly unavoidable consequence if $\varphi$ is quantifier-free and not 
``unreasonably strong''. A motivation of this claim follows
below after some examples.
\begin{enumerate}
\item First, if $\nu$ equals the maximal arity of a symbol in $\sigma$ and $\varphi(x_1, \ldots, x_\nu)$
expresses that $\neg R(x_1, \ldots, x_{\mr{ar}(R)})$ for every $R \in \sigma$,
then $\big|\mbY_n^0(\varphi)\big| = 1$ for all $n$ so the condition fails.

\item Let $\varphi(x, y) := (P(x) \wedge R(x, y)) \rightarrow P(y)$. 
Let the interpretation of $P$ be the whole domain $[n]$
(so all elements satisfy $P$). Then the interpretation of $R$ can be
chosen in $2^{n^2}$ ways (or $2^{\binom{n}{2}}$ if we require that $R$ is
symmetric and irreflexive) and every structure thus obtained will satisfy $\forall x, y \varphi(x, y)$.
Hence $\big|\mbY_n^0(\varphi)\big| \geq 2^{n^2}$ (or $\big|\mbY_n^0(\varphi)\big| \geq 2^{\binom{n}{2}}$
if $R$ is symmetric and irreflexive).

\item Let $\varphi(x, y, z) := (R(x, y) \wedge R(y, z)) \rightarrow R(x, z)$.
Divide $[n]$ into a ``left'' part of size $\lfloor n/2 \rfloor$ and a ``right'' part of size $\lceil n/2 \rceil$.
Interpret $R$ in such a way that if $R(a, b)$ holds then $a$ belongs to the left part and $b$ to the right part.
There are $2^{\lfloor n/2 \rfloor \cdot \lceil n/2 \rceil}$ such interpretations and all of them satisfy 
$\forall x, y, z \varphi(x, y, z)$. Hence $\big|\mbY_n^0(\varphi)\big| \geq 2^n$ for all sufficiently large $n$.

\item Let $\nu \geq 3$ and let $\varphi(x_1, \ldots, x_\nu) := \neg\bigwedge_{1 \leq i < j \leq \nu} R(x_i, x_j)$.
It follows from \cite{EKR} that $\big|\mbY_n^0(\varphi)\big| \geq 2^{\frac{n^2}{2}(1 - \frac{1}{\nu - 1}) + o(n^2)}$
so the condition is satisfied.

\item Let $\nu \geq 2$ and let 
\[
\varphi(x_1, \ldots, x_{\nu + 2}) := 
\bigg( \bigwedge_{2 \leq i < j \leq \nu + 2} x_i \neq x_j \bigg) \longrightarrow
\neg \bigwedge_{i = 2}^{\nu + 2} R(x_1, x_i).
\] 
Then $\mbY_n^0(\varphi)$ consists of all 
$\sigma$-structures such that every $a \in [n]$ is connected, via $R$, to at most $\nu$ other members of $[n]$.
It is well known that the number of $\nu$-regular graphs with domain $[n]$ is, for some $c$,
more than $2^{cn}$ for all large enough $n$ (see e.g. \cite{Bon}) and therefore 
$\big|\mbY_n^0(\varphi)\big| \geq 2^{cn}$.
\end{enumerate}
In more generality, 
suppose that $\varphi(x_1, \ldots, x_\nu)$ is quantifier-free.
Then $\bigcup_{n=1}^\infty \mbY_n^0$ is a {\em hereditary class} of structures in the sense that
if $\mcA \in \mbY_n^0$ and $\mcB$ is a substructure of $\mcA$, then $\mcB$ is isomorphic to a member of $\mbY_m^0$
for some $m \leq n$.
Suppose that the language $\sigma$ contains at least one relation symbol of arity at least 2 and that, for every $c > 0$
and every $N \in \mbbN$,
there is $n \geq N$ such that $\big|\mbY_n^0\big| < 2^{cn}$. 
It follows from results in \cite{LT} (building on a sequence of previous studies
of the ``asymptotic enumeration of hereditary classes'') that there is a polynomial $f(x)$ such that
$\big|\mbY_n^0\big| < f(n)$ for all sufficiently large $n$, and that $\bigcup_{n=1}^\infty \mbY_n^0$
is what the authors of \cite{LT} call a {\em basic} class. The definition of `basic class' is too technical to 
repeat here, but it implies, informally speaking, that all structures in $\bigcup_{n=1}^\infty \mbY_n^0$ are composed in a very
regular way, not just locally but also globally, which excludes even small amounts of ``randomness''.
This means, again informally speaking, that if the soft constraint
$\varphi(x_1, \ldots, x_\nu)$ is {\em quantifier-free} and  the ``axiom'' 
$\forall x_1, \ldots, x_\nu \varphi(x_1, \ldots, x_\nu)$ does not force all possible worlds in $\mbY_n^0$,
for all sufficiently large $n$, to be very regularly composed, thus omitting any randomness,
then condition~(\ref{the soft constraint is not too strong}) is satisfied.
}\end{rem}

\noindent
{\em In the rest of this section,
\begin{align*}
&\text{let $r \geq 1$ be the number of relation symbols in $\sigma$,}\\
&\text{let $\rho \geq 1$ be the maximal arity of a relation symbol in $\sigma$, and}\\
&f_w(n) := \frac{r n^\rho}{w} \ \text{ if $w > 0$}.
\end{align*}
}
Note that if $\nu > \rho$, then $\frac{f_w(n)}{n^\nu} \to 0$ as $n \to \infty$ so $f_w(n)$ is small compared
to the number of all $\nu$-tuples of elements from $[n]$ if $n$ is large.
The next theorem shows that
if an MLN $\mbbM$ has only one soft constraint with positive weight,
and the soft constraint satisfies some mild conditions, then,
using $\mbbP_n^\mbbM$ and assuming $n$ is large, the probability that a random sample 
$\mcA \in \mbW_n$ will have only few violations of the soft constraint is close to 1,
while if we use $\mbbP_n^{uni}$, the uniform distribution, the same probability is close to 0.

\begin{theor}\label{with high probability there are few violations}
Let $\mbbM = \big\{(\varphi(x_1, \ldots, x_\nu), w) \big\}$ where $w > 0$ be an MLN.
Suppose that, for all sufficiently large $n$,
$|\mbY_n^0(\varphi)| \geq 2^{cn}$ for some $c > 0$ that depends only on $\varphi$.\\
(a) For all sufficiently large $n$, $\mbbP_n^\mbbM\big(\mbY_n^{f_w(n)}(\varphi)\big) \geq 1 - 2^{-cn}$.\\
(b) If $\nu > \rho$ and each one of $\varphi(x_1, \ldots, x_\nu)  \wedge \mr{diff}(x_1, \ldots, x_\nu)$ and
$\neg\varphi(x_1, \ldots, x_\nu)  \wedge \mr{diff}(x_1, \ldots, x_\nu)$ is asymptotically persistent, then
there is $d > 0$ such that for all sufficiently large $n$,
$\mbbP_n^{uni}\big(\mbY_n^{f_w(n)}(\varphi)\big) \leq 2^{-dn}$
where $\mbbP_n^{uni}$ denotes the uniform distribution on $\mbW_n$. 
In particular we have
\[
\lim_{n\to\infty} \mbbP_n^\mbbM\big(\mbY_n^{f_w(n)}\big) = 1 \ \ 
\text{ and } \ \ \lim_{n\to\infty} \mbbP_n^{uni}\big(\mbY_n^{f_w(n)}\big) = 0.
\]
\end{theor}

\noindent
{\bf Proof.}
Let $\mbbM = \big\{(\varphi(x_1, \ldots, x_\nu), w) \big\}$ where $w > 0$ be an MLN
and let $\mbbP_n^\mbbM$ be the distribution on $\mbW_n$ determined by $\mbbM$.
Suppose that, for all sufficiently large $n$,
$|\mbY_n^0(\varphi)| \geq 2^{cn}$ for some $c > 0$ that depends only on $\varphi$ and
define $f_w(n) := \frac{r n^\rho}{w}$.
Then, for all sufficiently large $n$, we have 
\begin{align*}
&\mbbP_n^\mbbM\big(\mbW_n \setminus \mbY_n^{f_w(n)}(\varphi)\big) 
= \frac{\mu_n^\mbbM\big(\mbW_n \setminus \mbY_n^{f_w(n)}(\varphi)\big)}{\mu_n^\mbbM\big(\mbW_n\big)} 
\leq 
\frac{\mu_n^\mbbM\big(\mbW_n \setminus \mbY_n^{f_w(n)}(\varphi)\big)}{\mu_n^\mbbM\big(\mbY_n^0(\varphi)\big)} \\
&\leq \frac{\big|\mbW_n\big| \cdot 2^{w(n^\nu - f_w(n))}}{\big|\mbY_n^0(\varphi)\big| \cdot 2^{wn^\nu}} 
\leq \frac{2^{rn^\rho + wn^\nu - wf_w(n)}}{2^{cn + wn^\nu}}
= \frac{2^{rn^\rho + wn^\nu - rn^\rho}}{2^{cn + wn^\nu}}
= \frac{1}{2^{cn}}.
\end{align*}
Hence, for large enough $n$, $\mbbP_n^\mbbM\big(\mbY_n^{f_w(n)}(\varphi)\big) \geq 1 - 2^{-cn}$.

Suppose that $\nu > \rho$ and that each one of 
\[
\text{$\varphi(x_1, \ldots, x_\nu)$  $\wedge$ $\mr{diff}(x_1, \ldots, x_\nu)$ and
$\neg\varphi(x_1, \ldots, x_\nu)$ $\wedge$ $\mr{diff}(x_1, \ldots, x_\nu)$}
\]
 is asymptotically persistent.
It follows from 
Theorem~\ref{typical proportion of tuples satisfying a formula}
that there is $0 < \alpha < 1$ such that for every $\varepsilon > 0$ there is $d > 0$ such that for all sufficiently large $n$,
\begin{equation}\label{convergence of phi another time}
\frac{\big| \mbX_n^{\alpha, \varepsilon}(\varphi) \big|}{\big|\mbW_n\big|}
\ \geq \ 1 - e^{-dn}.
\end{equation}
Note that $[n]^\nu = \varphi(\mcA) \cup \neg\varphi(\mcA)$ where the union is disjoint.
Hence $n^\nu = |\varphi(\mcA)| + |\neg\varphi(\mcA)|$.
Observe that, since $\nu > \rho$, it follows that if $\varepsilon > 0$ then for all sufficiently large $n$, 
$f_w(n) = \frac{rn^\rho}{w} < \varepsilon n^\nu$.
If $\mcA \in \mbY_n^{f_w(n)}(\varphi)$ then $|\neg\varphi(\mcA)| \leq f_w(n)$,
hence $|\varphi(\mcA)| \geq n^\nu - f_w(n)$, so if $0 < \varepsilon < 1$ then for all sufficiently large $n$ we have
$|\varphi(\mcA)| \geq n^\nu - f_w(n)$
$= n^\nu - \frac{rn^\rho}{w}$ $> n^\nu - \varepsilon n^\nu$ $= (1 - \varepsilon)n^\nu$.
So if $0 < \varepsilon < 1$ is sufficiently small and $n$ is sufficiently large, 
\[
\mbY_n^{f_w(n)}(\varphi) \cap \mbX_n^{\alpha, \varepsilon}(\varphi) = \es.
\]
This together with (\ref{convergence of phi another time}) implies that, for some $d > 0$ and all sufficiently large $n$,
$\big|\mbY_n^{f_w(n)}(\varphi)\big| \Big/ \big|\mbW_n\big| \leq 2^{-dn}$.
\hfill $\square$

\medskip

\noindent
We now start working towards finding a kind of ``normal form'' for quantifier-free MLNs which
will be used to prove the main results of the next section. 
It may also be useful for proving future results about quantifier-free MLNs.

\begin{algorithm}\label{algorithm 1}{\rm
Let $\sigma$ be a language. The following algorithm converts any quantifier-free MLN $\mbbM$ over $\sigma$ into
an MLN $\mbbM'$ over $\sigma$ such that $\mbbP_n^{\mbbM'} = \mbbP_n^\mbbM$ for all $n \in \mbbN^+$
and every soft constraint of $\mbbM'$ is a maximal consistent conjunction of literals:
\begin{enumerate}
\item[] {\bf  Input:} a quantifier-free MLN $\mbbM$ over $\sigma$.

\item[] {\bf Step 1:} Let $\mbbM_1$ consist of all pairs $(\theta(\bar{x}), w)$ that satisfy the following conditions:
\begin{enumerate}
\item $\theta(\bar{x})$ is a maximal consistent conjunction of literals using variables from $\bar{x}$

\item there is some $(\varphi(\bar{x}), v) \in \mbbM$ such that $\theta(\bar{x})$ implies $\varphi(\bar{x})$

\item $w$ is the sum of all $v$ such that there is $(\varphi(\bar{x}), v) \in \mbbM$ such that $\theta(\bar{x})$ implies $\varphi(\bar{x})$.
\end{enumerate}

\item[] {\bf Step 2:} Let $\mbbM_2$ be obtained from $\mbbM_1$ by keeping (in $\mbbM_2$) only one of two pairs
$(\theta(\bar{x}), w)$ and $(\theta'(\bar{x}), w')$ if $\theta(\bar{x})$ and $\theta'(\bar{x})$ are equivalent
(and note that in this case $w = w'$, because of the construction of $\mbbM_1$).

\item[] {\bf Output:} $\mbbM' := \mbbM_2$.
\end{enumerate}
}\end{algorithm}

\noindent
It is straightforward to verify that, indeed, $\mbbP_n^{\mbbM'} = \mbbP_n^\mbbM$ for all $n \in \mbbN^+$.
{\em For the rest of this section we suppose that we have fixed some (linear) order on the countably
infinite set of all (logical object) variables.}

\begin{defin}\label{definition of variable reduction}{\rm
A {\em variable reduction} of a maximal consistent conjunction of $\sigma$-literals $\theta(\bar{x})$,
where $\bar{x} = (x_1, \ldots, x_k)$ is a sequence of distinct variables,
is a maximal consistent conjunction of $\sigma$-literals $\theta'(\bar{x}')$ such that 
$\rng(\bar{x}') \subseteq \rng(\bar{x})$,
$\theta'(\bar{x}')$ implies $x_i \neq x_j$ if $x_i, x_j \in \rng(\bar{x}')$ and $i \neq j$,
and for every finite $\sigma$-structure $\mcA$, $|\theta(\mcA)| = |\theta'(\mcA)|$.
}\end{defin}

\begin{algorithm}\label{algorithm for variable reduction}{\rm
The following algorithm takes a maximal consistent conjunction of literals and returns a
particular variable reduction of it which is determined by the order we fixed on all variables.
\begin{enumerate}
\item[] {\bf Input:} A maximal consistent conjunction of $\sigma$-literals $\theta(\bar{x})$.

\item[] {\bf Step 1:} For $x_i, x_j \in \rng(\bar{x})$ define $E(x_i, x_j)$ if and only if $\theta(\bar{x})$
implies $x_i = x_j$. Since $\theta(\bar{x})$ is maximal consistent it follows that $E$ is reflexive, 
symmetric, and transitive on $\rng(\bar{x})$, that is, $E$ is an equivalence relation on $\rng(\bar{x})$.

\item[] {\bf Step 2:} For every equivalence class $X$ of $E$, 
choose the first variable in $X$ according to the order on all variables, say $x_i$, and replace
every other (if there is any) variable in $\theta(\bar{x})$ which belongs to $X$ by $x_i$.

\item[] {\bf Output:} The maximal consistent conjunction of $\sigma$-literals obtained from Step~2.
\end{enumerate}
}\end{algorithm}

\noindent
It is straightforward to verify that the output of the algorithm is indeed a variable reduction of $\theta(\bar{x})$.
One may note that the particular ordering on $\bar{x}$ in Step~2 is irrelevant. The important
thing is to choose only one representative from every equivalence class of $E$.
Now we are ready to state and prove the ``normal form'' lemma.

\begin{lem}\label{normalized quantifier-free MLN}
Let $\mbbM$ be a quantifier-free MLN over $\sigma$.
Then there is a quantifier-free MLN $\mbbM'$ over $\sigma$ such that
\begin{itemize}
\item $\mbbM$ and $\mbbM'$ determine the same probability distribution on $\mbW_n$ for all $n$, and

\item for some $\nu \in \mbbN^+$ and $s_1, \ldots, s_\nu \in \mbbN$,
\[
\mbbM' = \bigcup_{k=1}^\nu 
\big\{ (\varphi_{k, 1}(x_1, \ldots, x_k), w_{k, 1}), \ldots, (\varphi_{k, s_k}(x_1, \ldots, x_k), w_{k, s_k})\big\}
\]
where, for all $k = 1, \ldots, \nu$ and $l = 1, \ldots, s_k$, $\varphi_{k, l}(x_1, \ldots, x_k)$ is a maximal consistent
conjunction of $\sigma$-literals which implies $x_i \neq x_j$ if $1 \leq i < j \leq k$.
\end{itemize}
\end{lem}

\noindent
{\bf Proof.}
Let $\mbbM$ be a quantifier-free MLN over $\sigma$.
We describe an algorithm for converting $\mbbM$ into a quantifier-free MLN $\mbbM'$ over $\sigma$
which satisfies the conditions of the lemma.

\begin{enumerate}
\item[] {\bf Step 1:} Run Algorithm~\ref{algorithm 1} on $\mbbM$ to get 
a quantifier-free MLN $\mbbM_2$ over $\sigma$ such that all of its soft constraints are
maximal consistent conjunctions of literals and $\mbbP_n^{\mbbM_2} = \mbbP_n^\mbbM$ for all $n$.

\item[] {\bf Step 2:} Let $\mbbM'$ consist of all pairs $(\theta'(\bar{x}'), w)$ such that there is
$(\theta(\bar{x}), v) \in \mbbM_2$ such that $\theta'(\bar{x}')$ is the variable reduction of $\theta(\bar{x})$
obtained by Algorithm~\ref{algorithm for variable reduction},
and $w$ is the sum of all $v$ such that there is $(\theta(\bar{x}), v) \in \mbbM_2$
such that $\theta'(\bar{x}')$ is the variable reduction of $\theta(\bar{x})$ obtained
by Algorithm~\ref{algorithm for variable reduction}.
\end{enumerate}
Then (as the reader can verify) $\mbbP_n^{\mbbM'} = \mbbP_n^{\mbbM_2}$ for all $n$,
and hence $\mbbP_n^{\mbbM'} =  \mbbP_n^\mbbM$ for all $n$.
By construction, $\mbbM'$ has the properties stated in the lemma.
\hfill $\square$

\section{Colour proportions expressible with quantifier-free MLNs
and the relative strength of MLNs versus lifted Bayesian networks}\label{Colour distributions expressible with quantifier-free MLNs}

\noindent
There are many different families of lifted probabilistic graphical models (LPGMs), 
for example MLNs, various kinds of lifted Bayesian networks,
and probabilistic logic programs. Therefore one may ask if every instance $\mbbG$ of 
an LPGM from one family can be replaced
by some instance $\mbbG'$ of an LPGM from another family in the sense that 
$\mbbG$ and $\mbbG'$ defined the same, or approximately the same, distribution on $\mbW_n$
for all sufficiently large domain sizes $n$. 
If the answer is no, then we really need different types of LPGMs in order to define different kinds
of probability distributions on possible worlds with large domain. 
If the answer is yes, then the difference between different types of LPGMs is just a matter of syntax, that is, the formal way
of specifying them.

In this section we will obtain negative answers with respect to a comparison between 
quantifier-free MLNs and lifted Bayesian networks. 
By lifted Bayesian networks we refer to a quite general family of LPGMs
which is defined below (Definitions~\ref{definition of LBN} and~\ref{semantics of LBN}).
Lifted Bayesian networks in this sense can define all probability distributions 
which can be defined by, for example, 
{\em relational Bayesian networks} \cite{Jae98a},
{\em Bayesian network specifications} \cite{CM}, {\em lifted Bayesian networks} in the sense of \cite{Kop20},
{\em $PLA$-networks} \cite{KW2}, or by 
{\em functional lifted Bayesian networks} \cite{Wei24}.
Informally speaking, Theorem~\ref{MLN with two convergence points}
says that there is an MLN $\mbbM$ over $\sigma := \{R\}$ where $R$ has arity 1 such that
there is {\em no} lifted Bayesian network $\mbbG$ such that for all sufficiently large $n$,
$\mbbP_n^\mbbG$ (the probability distribution on $\mbW_n$ determined by $\mbbG$)
approximates $\mbbP_n^\mbbM$ with as good accuracy as we like.
On the other hand, Theorem~\ref{limits of MLNs}
says that there is a lifted Bayesian network $\mbbG$ over $\sigma := \{R\}$ 
where $R$ has arity 1 such that
there is {\em no} quantifier-free MLN $\mbbM$ such that for all sufficiently large $n$,
$\mbbP_n^\mbbM = \mbbP_n^\mbbG$.
So over the language $\sigma = \{R\}$, where R has arity 1, quantifier-free MLNs and lifted Bayesian networks are
``incomparable'', that is, none of the families is more expressive than the other.

The language $\sigma = \{R\}$ where $R$ is unary is of course the simplest nontrivial language,
so one may ask if the situation changes for more complex and useful languages.
It turns out that we can use Theorems~\ref{MLN with two convergence points} and~\ref{limits of MLNs}
to prove a result, Theorem~\ref{MLN versus LBN generally},
which tells that quantifier-free MLNs over $\sigma$ and lifted Bayesian networks over $\sigma$
are incomparable 
for every language which contains at least one relation symbol that has arity 1.

We also prove (see Theorem~\ref{characterization of quantifier-free MLNs for colourings}),
in the setting where $\mbbM$ is a quantifier-free MLN over $\sigma = \{R\}$ and $R$ has arity 1, 
that, as the domain size tends to infinity, 
the proportion of elements that satisfy $R(x)$ concentrates (with high probabilty) around a finite set of reals
that depend only on $\mbbM$.
The proof, as well as examples below,  shows that the weights of the soft constraints of $\mbbM$ determine 
these reals.
Hence the weights do influence the asymptotic properties 
of $(\mbbP_n^\mbbM : n \in \mbbN^+)$.
This gives more input to the discussion of several researchers 
\cite{JBB, Mittal, Poole, Wei25}
regarding the influence of the weights of an MLN $\mbbM$ on $\mbbP_n^\mbbM$,
as $n \to \infty$. This is relevant for the question of whether the weights that have been
learned from ``example worlds'' with one domain are relevant for making inferences about possible
worlds with another domain of different size.
Theorem~\ref{characterization of quantifier-free MLNs for colourings}
also confirms the tendency that when there is a convergence phenomenon
for distributions determined 
by an LPGM as the domain size tends to infinity, then one gets good knowledge transfer properties and 
a fast method
for approximating probabilities of some (or many) events for arbitrarily large domains.
More is said about this after Theorem~\ref{characterization of quantifier-free MLNs for colourings}.

\subsection{Lifted Bayesian networks}\label{Lifted Bayesian networks}

In this section we define a notion of lifted Bayesian network (LBN) which generalizes many
(if not all) existing notions of ``lifted/\-parame\-trized/\-temp\-lated Bayesian network''
(see Remark~\ref{remark on other LBNs}).
We also prove a couple of basic lemmas about LBNs that will be used later.

\begin{defin}\label{definition of LBN}{\rm
Let $\sigma$ be a language.
A {\em lifted Bayesian network (LBN) over $\sigma$}, call it $\mbbG$, is determined by the following parts:
\begin{enumerate}
\item A {\em directed acyclic graph (DAG)}, also denoted $\mbbG$, with vertex set $\sigma$.
 For every $R \in \sigma$ let 
$\mr{par}_\mbbG(R)$ denote the set of parents of $R$ in $\mbbG$.

\item For each $R \in \sigma$ a map
\[
S_R : \big\{(\mcA, \bar{a}) : 
\text{  $\mcA$ is a finite $\mr{par}_\mbbG(R)$-structure and $\bar{a} \in A^{\mr{ar}(R)}$} \big\}
\to [0, 1]
\]
such that
\begin{enumerate}
\item [] if $f$ is an isomorphism from a finite $\mr{par}_\mbbG(R)$-structure $\mcA$ to a
finite $\mr{par}_\mbbG(R)$-structure $\mcB$, $\bar{a} \in A^{\mr{ar}(R)}$ and $f(\bar{a}) = \bar{b}$,
 then $S_R(\mcA, \bar{a}) = S_R(\mcB, \bar{b})$.
\end{enumerate}
\end{enumerate}
}\end{defin}

\noindent
The intuition behind part~(2) above is that if $R \in \sigma$ and $\bar{a} \in A^{\mr{ar}(R)}$
where $A$ is the domain of a $\sigma$-structure, then the probability that $R(\bar{a})$ holds should
depend only (or at most) on the reduct of the $\sigma$-structure to the parents of $R$ with respect to $\mbbG$.

\begin{defin}\label{semantics of LBN}{\rm
Let $\sigma$ be a language and $\mbbG$ an LBN over $\sigma$ and let (as usual)
$\mbW_n$ be the set of $\sigma$-structures with domain $[n]$. 
For every $n \in \mbbN^+$
and every $\mcA \in \mbW_n$ let
\[
\mbbP_n^\mbbG(\mcA) = \prod_{R \in \sigma} 
\prod_{\bar{a} \in R^\mcA} S_R(\mcA \uhrc \mr{par}_\mbbG(R), \bar{a}) 
\prod_{\substack{\bar{a} \in [n]^{\mr{ar}(R)} \\ \bar{a} \notin R^\mcA}} 
\big(1 - S_R(\mcA \uhrc \mr{par}_\mbbG(R), \bar{a}) \big).
\]
For $\mbX \subseteq \mbW_n$, let $\mbbP_n^\mbbG(\mbX) = \sum_{\mcA \in \mbX} \mbbP_n^\mbbG(\mcA)$.
}\end{defin}

\noindent
Then $\mbbP_n^\mbbG$ is a probability distribution on $\mbW_n$ (we omit the tedious but
standard verification of this) which we call the {\em probability distribution (on $\mbW_n$) 
determined by $\mbbG$}.

\begin{exam}\label{example 1 of LBN}{\rm
Let $\sigma = \{S, C, F\}$ where we can, following earlier examples, think of $S(x)$, $C(x)$, and $F(x, y)$, 
as meaning ``$x$ smokes'', ``$x$ has cancer'', and ``$x$ and $y$ are friends'', respectively.
Then an LBN over $\sigma$ is specified by the following DAG and the following specification of the maps
$S_S$, $S_C$, and $S_F$:
\begin{center}
\includegraphics[scale=1.1]{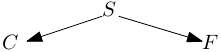}
\end{center}
\begin{itemize}
\item For every finite $\es$-structure $\mcA$ and $a \in A$, 
$S_S(\mcA, a) = 0.1$.
In words, ``with probability $0.1$ a person smokes, 
independently of what other people do''.

\item For every finite $\{S\}$-structure $\mcA$ and $a \in A$, 
$S_C(\mcA, a) = \begin{cases} 0.25 \text{ if $\mcA \models S(a)$}, \\
0.03 \text{ if $\mcA \models \neg S(a)$}.
\end{cases}$.
In words, ``if $a$ is a smoker then the probability that $a$ has cancer is $0.25$,
independently of what the case is for other smokers, and otherwise the
probability that $a$ has cancer is $0.03$''.

\item For every finite $\{S\}$-structure $\mcA$ and $(a, b) \in A^2$, \\
$S_F(\mcA, (a, b)) = 
\begin{cases} 0.2 \text{ if }\mcA \models S(a) \leftrightarrow S(b) \\
0.1 \text{ if }\mcA \models S(a) \leftrightarrow \neg S(b).
\end{cases}$
In words, ``if $a$ and $b$ are smokers, or none of them is a smoker,
the probability that they are friends is $0.2$, and
otherwise the probability that they are friends is $0.1$''.
\end{itemize}
}\end{exam}

\begin{exam}\label{example 2 of LBN}{\rm
This example illustrates that LBNs can specify probabilities by aggregating over the domain.
Let $\sigma = \{Cr, P, T\}$ where we can think of $Cr(x)$, $P(x)$, and $T(x)$ as
meaning ``$x$ is corrupt'', ``$x$ is a politician'', and ``$x$ has trust in society'', respectively.
The LBN over $\sigma$ described below by the given DAG and definition of $S_R$
for $R \in \{Cr, P, T\}$ expresses, with informal language, 
that the probability of being corrupt is 0.1, 
the probability
of being a politician is $0.03$ (independently of what other persons are), 
and the probability that a person $x$ has trust in society is
\begin{enumerate}
\item[] 1 - (proportion of corrupt persons)$\cdot$ 0.6 if $x$ is politician, and
\item[] 1 - (proportion of corrupt persons)$\cdot$ 0.8 if $x$ is not a politician.
\end{enumerate}

\medskip

\begin{center}
\includegraphics[scale=1.1]{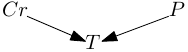}
\end{center}
\begin{itemize}
\item For every finite $\es$-structure $\mcA$ and $a \in A$, $S_{Cr}(\mcA, a) = 0.1$,
and $S_P(\mcA, a) = 0.03$.

\item For every finite $\{Cr, P\}$-structure $\mcA$ and $a \in A$,\\
$S_T(\mcA, a) = 
\begin{cases} 1 - 0.6 \frac{|Cr(\mcA)|}{|A|} \text{ if }\mcA \models P(a),\\
1 - 0.8 \frac{|Cr(\mcA)|}{|A|} \text{ if }\mcA \models \neg P(a).
\end{cases}$
\end{itemize}
A grounding of this LBN on, say, the domain $D = \{a, b, c\}$ is obtained by first letting the vertex set
of the grounded network consist of all groundings of the atomic formulas 
$Cr(x)$, $P(x)$, and $T(x)$ on $D$. For every pair of grounded atomic formulas, say
$Cr(b)$ and $T(a)$ we draw a directed edge from $Cr(b)$ to $T(a)$ if there is an edge
from $Cr$ to $T$ in $\mbbG$ (which is the case) and the value of $S_T(\mcA, a)$ 
is influenced by the truth value of $Cr(b)$ in $\mcA$
as $\mcA$ ranges over $\{Cr, P\}$-structures with domain $D$.
In the example the value of $S_T(\mcA, a)$ is influenced by the truth values of $Cr(a)$, $Cr(b)$ and $Cr(c)$,
so the grounded network will have edges from $Cr(a)$, $Cr(b)$ and $Cr(c)$ to $T(a)$.
However, there will not be an edge from $P(b)$ to $T(a)$, but there will be one from $P(a)$ to $T(a)$.
By reasoning according the principle that was informally described we get the following grounded network:
\begin{center}
\includegraphics[scale=1.1]{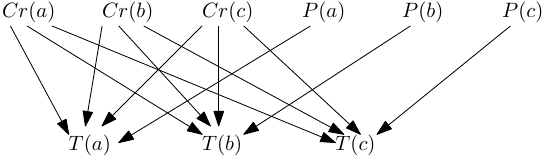}
\end{center}
}\end{exam}

\begin{rem}\label{remark on other LBNs}{\rm
Every probability distribution $\mbbP_n$ on $\mbW_n$ (for any $n$) which can be defined
by some {\em relational Bayesian network} \cite{Jae98a},
{\em Bayesian network specification} \cite{CM}, {\em lifted Bayesian network} in the sense of \cite{Kop20},
{\em $PLA$-network} \cite{KW2}, or by some
{\em functional lifted Bayesian network} \cite{Wei24} can be defined by some  LBN in the sense 
of Definition~\ref{definition of LBN} above. 
In fact, the specification of a relational Bayesian network, Bayesian network specification, lifted Bayesian network,
$PLA$-network, or functional lifted Bayesian network, can be translated fairly straightforwardly into an LBN in the sense of
Definition~\ref{definition of LBN} which determines the same probability distribution on $\mbW_n$ for all $n$.
In the mentioned cases the map $S_R$ (from Definition~\ref{definition of LBN}) is determined by 
one or more logical formulas from some (possibly many valued) formal logic
used by the respective formalism in \cite{Jae98a, CM, Kop20, KW2, Wei24}.
Since the (truth) values of the formulas considered are preserved by isomorphisms, 
part~(2) of Definition~\ref{definition of LBN} follows.
}\end{rem}

\noindent
Suppose that $\mbbG$ is an LBN over $\sigma$ and $Q, R \in \sigma$.
Furthermore, suppose that there is an edge from $Q$ to $R$ in the DAG associated to $\mbbG$.
It is possible that $S_R$ (from Definition~\ref{definition of LBN}) is indifferent to the 
interpretation of $Q$ in every finite $\mr{par}_\mbbG(R)$-structure, and in this case the edge
from $Q$ to $R$ does not reflect a true dependency, so we can view this edge as being redundant.
We make this intuition precise in the definition below and then we show that all redundant edges
can be removed without changing
the probability distribution that $\mbbG$ determines on $\mbW_n$ for all sufficiently large $n$.

\begin{defin}\label{definition of redundant edge}{\rm
Let $\mbbG$ be an LBN over $\sigma$, let $R, Q \in \sigma$ and suppose that $Q \in \mr{par}_\mbbG(R)$,
so there is an edge from $Q$ to $R$.
We say that the edge from $Q$ to $R$ is {\em asymptotically redundant} if the following holds:
There is $n_0 \in \mbbN^+$ such that whenever $\mcA$ is a finite 
$(\mr{par}_\mbbG(R) \setminus \{Q\})$-structure, $|A| \geq n_0$,
$\bar{a} \in A^{\mr{ar}(R)}$, and $\mcA_1$ and $\mcA_2$ are expansions of $\mcA$ to 
$\mr{par}_\mbbG(R)$, then $S_R(\mcA_1, \bar{a}) = S_R(\mcA_2, \bar{a})$.
}\end{defin}

\begin{lem}\label{removing redundant edges}
Let $\mbbG$ be an LBN over $\sigma$.
Then there is an LBN $\mbbG^*$ over $\sigma$ without any asymptotically redundant edge and such that
$\mbbP_n^{\mbbG^*} = \mbbP_n^\mbbG$ for all sufficiently large $n \in \mbbN^+$.
\end{lem}

\noindent
{\bf Proof.}
It suffices to prove the following claim:
\begin{enumerate}
\item[] Let $\mbbG$ be an LBN over $\sigma$ and suppose that $R \in \sigma$, $Q \in \mr{par}_\mbbG(R)$
and the edge from $Q$ to $R$ is asymptotically redundant. 
Then there is an LBN $\mbbG'$ over $\sigma$ such that 
$\mbbP_n^{\mbbG'} = \mbbP_n^\mbbG$ for all sufficiently large $n \in \mbbN^+$, and
the DAG of $\mbbG'$ is obtained
from the DAG of $\mbbG$ by only removing the edge from $Q$ to $R$ and modifying $S_R$.
\end{enumerate}
Let $\mbbG$ be as assumed, so there is an edge from $Q$ to $R$ in $\mbbG$ which is asymptotically redundant.
Then there is $n_0$ such that whenever $\mcA$ is a finite 
$(\mr{par}_\mbbG(R) \setminus \{Q\})$-structure, $|A| \geq n_0$,
$\bar{a} \in A^{\mr{ar}(R)}$, and $\mcA_1$ and $\mcA_2$ are expansions of $\mcA$ to 
$\mr{par}_\mbbG(R)$, then $S_R(\mcA_1, \bar{a}) = S_R(\mcA_2, \bar{a})$.

For every finite $(\mr{par}_\mbbG(R) \setminus \{Q\})$-structure $\mcA$,
let $\hat{\mcA}$ be the expansion of $\mcA$ to $\mr{par}_\mbbG(R)$ such
that $Q^{\hat{\mcA}} = \es$, and then define $S'_R(\mcA, \bar{a}) = S_R(\hat{\mcA}, \bar{a})$.
It then follows that if $\mcA$ and $\mcB$ are finite $(\mr{par}_\mbbG(R) \setminus \{Q\})$-structures,
$\bar{a} \in A^{\mr{ar}(R)}$, $\bar{b} \in B^{\mr{ar}(R)}$,
and $f$ is an isomorphism from $\mcA$ to $\mcB$ such that $f(\bar{a}) = \bar{b}$, then $f$ is also
an isomorphism from $\hat{\mcA}$ to $\hat{\mcB}$, so (by the definition of an LBN applied to $\mbbG$)
\begin{equation}\label{S'-R and S'-R}
S'_R(\mcA, \bar{a}) = S_R(\hat{\mcA}, \bar{a}) = S_R(\hat{\mcB}, \bar{b}) = S'_R(\mcB, \bar{b}).
\end{equation}
Let $\mbbG'$ be the LBN over $\sigma$ which is obtained from $\mbbG$ by removing the
edge in $\mbbG$ from $Q$ to $R$ and by replacing $S_R$ by $S'_R$.
By~(\ref{S'-R and S'-R}), $\mbbG'$ is indeed an LBN.

By assumption, for any  finite $(\mr{par}_\mbbG(R) \setminus \{Q\})$-structure 
$\mcA$ with cardinality at least $n_0$
and any expansion $\tilde{\mcA}$ of $\mcA$ to $\mr{par}_\mbbG(R)$ we have
$S_R(\tilde{\mcA}, \bar{a}) = S_R(\hat{\mcA}, \bar{a}) = S'_R(\mcA, \bar{a})$.
Therefore we have $\mbbP_n^{\mbbG'} = \mbbP_n^\mbbG$ for all $n \geq n_0$.
\hfill $\square$

\medskip

\noindent
We will use the following basic result (which can be generalized to higher arities):

\begin{lem}\label{independency in LBNs}
Let $\mbbG$ be an LBN over $\sigma$ and suppose that $R \in \sigma$ 
has arity 1 and has no parent
in the DAG associated to $\mbbG$.
Fix any $n \in \mbbN^+$ and let $a_1, \ldots, a_m \in [n]$ be distinct.
Then, for all $1 \leq i < j \leq m$,
 $\mbbP_n^\mbbG\big(\{\mcA \in \mbW_n : \mcA \models R(a_i)\}\big) = 
\mbbP_n^\mbbG\big(\{\mcA \in \mbW_n : \mcA \models R(a_j)\}\big)$
and the event $\{\mcA \in \mbW_n : \mcA \models R(a_1)\}$ is independent from all events
$\{\mcA \in \mbW_n : \mcA \models R(a_i)\}$, $i = 2, \ldots, m$.
\end{lem}

\noindent
{\bf Proof.}
Let $\mbbG$ and $R \in \sigma$ be as assumed. Fix any $n$.
Since $R$ has no parent we get $\mr{par}_\mbbG(R) = \es$, so for every $\mcA \in \mbW_n$,
$\mcA \uhrc \mr{par}_\mbbG(R) = \mcA \uhrc \es$. 
Since a $\es$-structure is just a set with no further structure
and $\mcA$ has domain $[n]$ it follows that $\mcA \uhrc \mr{par}_\mbbG(R)$
is just the set $[n]$. For any $a, b \in [n]$ there is a permutation (bijection) $f : [n] \to [n]$
such that $f(a) = b$ and this $f$ is an isomorphism of $[n]$ viewed as a $\es$-structure.
By part~(2) of Definition~\ref{definition of LBN} it follows that for all $\mcA, \mcB \in \mbW_n$
and all $a, b \in [n]$ we have
$S_R(\mcA \uhrc \mr{par}_\mbbG(R), a) = S_R(\mcB \uhrc \mr{par}_\mbbG(R), b)$.

From Definition~\ref{semantics of LBN} we get, for all $i = 1, \ldots, n$,
\begin{align}\label{probability of R(a) equals S-R}
&\mbbP_n^\mbbG\big(\{\mcA \in \mbW_n : \mcA \models R(a_i)\}\big) = 
S_R(\mcA \uhrc \mr{par}_\mbbG(R), a_i) \quad \text{ and } \\
&\mbbP_n^\mbbG\big(\{\mcA \in \mbW_n : \mcA \models R(a_j)\}\big) = 
S_R(\mcA \uhrc \mr{par}_\mbbG(R), a_j). \nonumber
\end{align}
As $S_R(\mcA \uhrc \mr{par}_\mbbG(R), a_i) = S_R(\mcA \uhrc \mr{par}_\mbbG(R), a_j)$
the first claim of the lemma follows.
By Definition~\ref{semantics of LBN} again, we have
\[
\mbbP_n^\mbbG\big(\{\mcA \in \mbW_n : \mcA \models \bigwedge_{i=1}^m R(a_i)\}\big) = 
\prod_{i=1}^m S_R(\mcA \uhrc \mr{par}_\mbbG(R), a_i),
\]
so the independence claim of the lemma follows from~(\ref{probability of R(a) equals S-R}).
\hfill $\square$

\subsection{Quantifier free MLNs over a language with only one relation symbol}
\label{Quantifier free MLNs over a signature with only one relation symbol}

In this section we consider the simplest case of a nonempty language $\sigma$, namely when 
$\sigma = \{R\}$ where $R$ is a unary relation symbol, and we can think of $R(x)$ as meaning ``$x$ is coloured''.
We show that if $\mbbM$ is a quantifier-free MLN over $\sigma$ then there are finitely many
reals such that, for all $\varepsilon > 0$, if $n$ is large enough then, with probability $\geq 1 - \varepsilon$,
the proportion of coloured elements is within distance $\varepsilon$ from one of these reals
(Theorem~\ref{characterization of quantifier-free MLNs for colourings}).
We also show that quantifier-free MLNs over $\sigma$ and LBNs over $\sigma$ are incomparable 
(Theorems~\ref{MLN with two convergence points}
and~\ref{limits of MLNs}).
A similar result for more general languages, which builds on the results of this section,
is given in Section~\ref{MLNs versus LBNs over richer languages}.

We will use the following notation, where $0 \leq m \leq n$ are integers:
\[
\mbW_n^m := \big\{\mcA \in \mbW_n : |R(\mcA)| = m\}.
\]
So intuitively speaking, $\mbW_n^m$ consists of the structures from $\mbW_n$ in which exactly $m$ elements are coloured,
and note that $\mbW_n^0$ and $\mbW_n^n$ are singleton sets.

The next result states, in a precise way, that there is an MLN $\mbbM$ over $\sigma$ such that no LBN 
over $\sigma$ can approximate
the distributions $\mbbP_n^\mbbM$, for large $n \in \mbbN^+$, that are determined by $\mbbM$,
let alone define them exactly.

\begin{theor}\label{MLN with two convergence points}
Let 
$\mbbM = \big\{ (\varphi_0(x, y), w_0), \ (\varphi_1(x, y), w_1), \ (\varphi_2(x, y), w_2) \big\}$
be an MLN where $w_0 = w_2 = 1$, $w_1 = 0$,
\begin{enumerate}
\item[] $\varphi_0(x, y)$ is the formula $\neg R(x) \wedge \neg R(y)$,
\item[] $\varphi_1(x, y)$ is the formula $(R(x) \wedge \neg R(y)) \vee (\neg R(x) \wedge R(y))$, and
\item[] $\varphi_2(x, y)$ is the formula $R(x) \wedge R(y)$.
\end{enumerate}
Then $\lim_{n\to\infty} \mbbP_n^\mbbM\big(\mbW_n^0\big) = \lim_{n\to\infty} \mbbP_n^\mbbM\big(\mbW_n^n\big) = \frac{1}{2}$,
and hence, 
for all large enough $n$ and all distinct $i, j \in [n]$, 
the event that $R(i)$ holds is {\rm dependent} on the event that $R(j)$ holds.

Let $\mbbG$ be an LBN over $\sigma$.
For all large enough $n$ there is $\mbX_n \subseteq \mbW_n$ such that 
$\big|\mbbP_n^\mbbG(\mbX_n) - \mbbP_n^\mbbM(\mbX_n)\big| \geq 0.01$.
\end{theor}

\noindent
The technical proof of 
Theorem~\ref{MLN with two convergence points} is given in 
Appendix~\ref{proof of MLN with two convergence points},
but the next remark hints at why the theorem holds.

\begin{rem}\label{remark about MLN with two convergence points}{\rm
 (Concretization of Theorem~\ref{MLN with two convergence points})
Let $\mbbM$ be the MLN from Theorem~\ref{MLN with two convergence points}.
For any $0 \leq m \leq n$ and $\mcA \in \mbW_n^m$ we have
\begin{align*}
\mu_n^\mbbM(\mcA) = 2^{|\varphi_0(\mcA)| + |\varphi_2(\mcA)|} = 
2^{|R(\mcA)|^2 + |\neg R(\mcA)|^2} = 2^{m^2 + (n-m)^2}.
\end{align*}
As a subset of $[n]$ of size $m$ can be chosen in $\binom{n}{m}$ ways we get
\[
\mu_n^\mbbM(\mbW_n^m) = \binom{n}{m} 2^{m^2 + (n-m)^2}.
\]
Since $\binom{n}{n-m} = \binom{n}{m}$ it follows that 
$\mu_n^\mbbM(\mbW_n^m) = \mu_n^\mbbM(\mbW_n^{n-m})$ for all $0 \leq m \leq n$.
Let us compute $\mbbP_n^\mbbM(\mbW_n^0 \cup \mbW_n^n)$ for $n = 4, 5$ which will show that
already for these small $n$ the probability is close to 1, and for $n = 5$ it is considerably closer to 1 than for $n = 4$.
We get 
$\mu_4^\mbbM(\mbW_4^0) = \mu_4^\mbbM(\mbW_4^4) = 2^{16}$,
$\mu_4^\mbbM(\mbW_4^1) = \mu_4^\mbbM(\mbW_4^3) = 4 \cdot 2^{10}$, and
$\mu_4^\mbbM(\mbW_4^2) = 6 \cdot 2^8$.
Therefore 
\begin{align*}
\mbbP_4^\mbbM(\mbW_4^0 \cup \mbW_4^4) = 
\frac{2^{17}}{2^{17} + 8 \cdot 2^{10} + 6 \cdot 2^{8}} = \frac{256}{275} \approx 0.9309.
\end{align*}
By similar calculations we get
\[
\mbbP_5^\mbbM(\mbW_5^0 \cup \mbW_5^5) = \frac{2048}{2093} \approx 0.9785.
\]
So $\mbbP_5^\mbbM(\mbW_5^0) = \mbbP_5^\mbbM(\mbW_5^5) = \frac{2048}{2 \cdot 2093} \approx 0.4892$.
If $\mbbP$ is a probability distribution on $\mbW_5$ and $\psi$ is a formula then let 
$\mbbP(\psi) := \mbbP(\{\mcA \in \mbW_5 : \mcA \models \psi \})$.
It follows that if $i, j \in [5]$ are distinct then $\mbbP_5^\mbbM(R(i))$ is close to $1/2$
and $\mbbP_5^\mbbM(R(i) \wedge R(j))$  is also close to $1/2$. 
Therefore the event that $R(i)$ holds is {\em not} independent from the event that $R(j)$ holds 
if we use $\mbbP_5^\mbbM$ (as the latter probability
must be close to $1/4$ if they are independent).

But due to Lemma~\ref{independency in LBNs},
if $\mbbG$ is an LBN over $\sigma$ then, with respect to $\mbbP_5^\mbbG$,
the event that $R(i)$ holds is independent from the event that $R(j)$ holds.
It follows that if $\mbbP_5^\mbbG(R(i)) \approx 1/2$ then 
$\mbbP_5^\mbbG(R(i) \wedge R(j)) \approx 1/4$, so
$|\mbbP_5^\mbbM(R(i) \wedge R(j)) - \mbbP_5^\mbbG(R(i) \wedge R(j))| \approx 1/4 \geq 0.01$.
(The bound 0.01 is a big understatement, but makes it easy to follow the final part of the rigorous proof in
Appendix~\ref{proof of MLN with two convergence points}.)
}\end{rem}

\noindent
Since, in this subsection, we only work with one relation symbol which has arity 1, which can represent a colour, 
the only interesting property of a 
structure is the number, or proportion, of coloured elements in it.
Informally speaking the next theorem, 
which is proved in Appendix~\ref{proof of characterization of quantifier-free MLNs for colourings},
states that if the sequence of probability distributions is determined by a quantifier-free MLN,
then, for large $n$, this proportion is almost surely approximately equal 
to one of the numbers in a finite set of numbers determined by the MLN.

\begin{theor}\label{characterization of quantifier-free MLNs for colourings}
Let $\mbbM$ be a quantifier-free MLN over $\sigma$.
Then there are $t \in \mbbN^+$ and $\alpha_1, \ldots, \alpha_t \in [0, 1]$, depending only on $\mbbM$,
such that for all $\varepsilon > 0$,
\[
\lim_{n\to\infty} \mbbP_n^\mbbM\Big( \Big\{ \mcA \in \mbW_n : 
\frac{|R(\mcA)|}{n} \in \bigcup_{i=1}^t (\alpha_i - \varepsilon, \alpha_i + \varepsilon) \Big\}\Big) = 1.
\]
If all soft constraints of $\mbbM$ with positive weights have only one variable then $t=1$ and $\alpha_1$
can be computed directly from the weights of the soft constraints of $\mbbM$. 
Otherwise the numbers $\alpha_1, \ldots, \alpha_t$ are the points in the interval $[0, 1]$ where a 
polynomial that depends only on $\mbbM$ reaches its maximum when restricted to $[0, 1]$.
In particular, these numbers depend on the weights of the soft constraints of $\mbbM$.
\end{theor}

\begin{exam}\label{example of colouring}{\rm
Consider the following MLN
\[
\mbbM := \big\{\big((R(x) \wedge \neg R(y)) \vee (\neg R(x) \wedge R(y)) , 2 \big), \ 
\big(\neg R(x) \wedge \neg R(y), 1 \big)\big\}.
\]
Each ordered pair $(a, b)$ such that exactly one of $a$ and $b$ is coloured contributes with 2
to the exponent in the definition of $\mu_n^\mbbM$.
Each ordered pair $(a, b)$ such that neither $a$ nor $b$ is coloured
contributes with 1 to the same exponent. Pairs where both $a$ and $b$ are coloured contribute nothing.
Intuitively speaking, $\mu_n^\mbbM$ gives the highest value to 
structures with both coloured and uncoloured elements but with
fewer coloured than uncoloured elements.
Suppose that $\mcA \in \mbW_n^m$, so exactly $m$ elements of $\mcA$ are coloured.
It follows that there
are $2m(n-m)$ ordered pairs such that exactly one of the members of the pair are coloured, and
there are $(n-m)^2$ ordered pairs such that none of the members of the pair is coloured.
So if $\mcA \in \mbW_n^m$ and $f(\alpha) := 4\alpha(1-\alpha) + (1-\alpha)^2$, then
\[
\mu_n^\mbbM(\mcA) = \exp_2\big(2 \cdot 2m(n-m) + (n-m)^2\big) = 
\exp_2\big(n^2 f(m/n)\big).
\]
Since a subset of $[n]$ of size exactly $m$ can be chosen in $\binom{n}{m}$ ways we get
\[
\mu_n^\mbbM\big(\mbW_n^m\big) = \binom{n}{m} \exp_2\big(n^2 f(m/n)\big).
\]
By considering the first and second derivatives of $f$ we see that $f(\alpha)$ reaches its unique maximum when
$\alpha = 1/3$. In this example, $f$ is the function that 
Theorem~\ref{characterization of quantifier-free MLNs for colourings}
mentions (which becomes clear from the details of the proof of that theorem).
So for large $n$ the probability is high that the proportion of
coloured elements is close to 1/3.
Already for small $n$, such as for example $n = 6$, this tendency is clear.
We have:
\begin{center}
\begin{tabular}{|l l l|}
\hline 
$\mu_6^\mbbM\big(\mbW_6^0\big) = 2^{36}$ &
$\mu_6^\mbbM\big(\mbW_6^1\big) = 6 \cdot 2^{45}$ &
$\mu_6^\mbbM\big(\mbW_6^2\big) = 15 \cdot 2^{48}$ \\
$\mu_6^\mbbM\big(\mbW_6^3\big) = 20 \cdot 2^{45}$ &
$\mu_6^\mbbM\big(\mbW_6^4\big) = 15 \cdot 2^{36}$ &
$\mu_6^\mbbM\big(\mbW_6^5\big) = 6 \cdot 2^{21}$ \\
$\mu_6^\mbbM\big(\mbW_6^6\big) = 1$ & & \\
\hline
\end{tabular}
\end{center}
This gives for example
\begin{align*}
\mbbP_6^\mbbM\big(\mbW_6^2\big) = 
\frac{\mu_6^\mbbM\big(\mbW_6^2\big)}{\mu_6^\mbbM\big(\mbW_6\big)} = 
\frac{\mu_6^\mbbM\big(\mbW_6^2\big)}
{\sum_{i=0}^6 \mu_6^\mbbM\big(\mbW_6^i\big)} 
= \frac{1 \ 407 \ 374 \ 883 \ 553 \ 280}{1 \  712 \ 672 \ 616 \ 393 \ 387} 
\approx 0.8217.
\end{align*}
As a contrast, the function associated to the MLN in
Theorem~\ref{MLN with two convergence points}
is $g(\alpha) = \alpha^2 + (1 - \alpha)^2$ which, when restricted to $[0, 1]$, reaches its maximum
when $\alpha = 0$ and when $\alpha = 1$.
}\end{exam}

\begin{exam}\label{degree 4 example}{\rm
This example illustrates that by using soft constraints with more variables we can get
asymptotic behaviours that cannot be obtained with fewer variables.
Let 
\[
\mbbM = \big\{\big(R(x_1) \wedge \bigwedge_{i=2}^5 \neg R(x_i), \ 2\big), \ 
\big(\neg R(x_1) \wedge \bigwedge_{i=2}^5 R(x_i), \ 2\big) \big\}.
\]
Then 
\begin{align*}
\mu_n^\mbbM\big(\mbW_n^m\big) &= \binom{n}{m} 2^{2m(n-m)^4 + 2(n-m)m^4}\\
&= \binom{n}{m} \exp_2\bigg(n^5 \bigg(2\bigg(\frac{m}{n}\bigg)\bigg(1 - \frac{m}{n}\bigg)^4 + 
2\bigg(1 - \frac{m}{n}\bigg)\bigg(\frac{m}{n}\bigg)^4 \bigg)\bigg),
\end{align*}
and the function associated to $\mbbM$ by 
Theorem~\ref{characterization of quantifier-free MLNs for colourings}
is 
\[
f(\alpha) = 2\alpha(1 - \alpha)^4 + 2(1 - \alpha)\alpha^4.
\]
It reaches its maximum $1/6$ when $\alpha = \frac{1}{2} + \frac{1}{2\sqrt{3}}$ and when
$\alpha = \frac{1}{2} - \frac{1}{2\sqrt{3}}$.
Due to the symmetry of $f$ around the vertical axis $\alpha = 1/2$ we can conclude that
for every $\varepsilon > 0$, if $n$ is large enough then the probability that the proportion of coloured
elements is within distance $\varepsilon$ from  $\frac{1}{2} + \frac{1}{2\sqrt{3}}$ is in $(1/2 - \varepsilon, 1/2 + \varepsilon)$,
and the same holds for $\frac{1}{2} - \frac{1}{2\sqrt{3}}$.
This behaviour is not possible to obtain by a quantifier-free MLN (over the same language) such that all soft constraints
have at most two variables, because then the associated polynomial has degree at most 2 
(as revealed by the proof of Theorem~\ref{characterization of quantifier-free MLNs for colourings}),
so if its restriction to $[0, 1]$ reaches
its maximum in two points then these points are 0 and 1.
}\end{exam}

\noindent
Theorems~\ref{characterization of quantifier-free MLNs for colourings} and
\ref{MLN with two convergence points}
and the examples
show that the weights of a quantifier-free MLN $\mbbM$ over $\sigma$ do influence the
asymptotic behaviour of $(\mbbP_n^\mbbM : n \in \mbbN^+)$.
Theorem~\ref{characterization of quantifier-free MLNs for colourings}
also confirms the tendency that probabilistic convergence phenomena often
imply good knowledge transfer properties and good scalability properties
(see e.g. \cite{Jae98a, CM, Kop20, KW1, KW2, Kop24, KT, Wei21, Wei24}).
A structure $\mcA \in \mbW_n$ is characterized by the proportion of elements of $[n]$ that are coloured.
From the soft constraints and weights of $\mbbM$ we can construct a differentiable function
$f(x)$ as done in the proof of Theorem~\ref{characterization of quantifier-free MLNs for colourings}
in Appendix~\ref{proof of characterization of quantifier-free MLNs for colourings}.
As seen from that proof, the (finitely many) points $\alpha_1, \ldots, \alpha_t \in [0, 1]$ such that 
$f(\alpha_i) = \sup\{f(\beta) : \beta \in [0, 1]\}$, for $i = 1, \ldots, t$, 
have the property that, for any $\varepsilon > 0$, 
with probability tending to 1 as
$n \to \infty$, the proportion of coloured elements of  a random $\mcA \in \mbW_n$ 
is in $\bigcup_{i=1}^t(\alpha_i - \varepsilon, \alpha_i + \varepsilon)$.
Hence, knowledge about the typical proportions of coloured elements transfer across all sufficiently
large domains.
The global maximizers on $[0, 1]$ can be computed (or at least estimated) from $f(x)$ by using only basic methods from calculus
(implemented in numerous softwares),
without any reference to the domain size $n$ or $\mbW_n$, so it is computationally very resource efficient.

The next result, proved in Appendix~\ref{proof of limits of MLNs},
is a converse of
Theorem~\ref{MLN with two convergence points} 
and shows that there are LBNs such that no quantifier-free MLN can define the same distribution on
the set of possible worlds with sufficiently large domain.

\begin{theor}\label{limits of MLNs}
Choose any real $\alpha > 0$. Let $\mbbG$ be the LBN over $\sigma = \{R\}$ where $R$ has the associated map
$S_R(\mcA, a) = n^{-\alpha}$ for every finite $\es$-structure $\mcA$ with domain size $n$ and
every member $a$ of the domain of $\mcA$.
Then, for all $n \in \mbbN^+$ and $\mcA \in \mbW_n$,
\[
\mbbP_n^\mbbG(\mcA) = p_n^{|R(\mcA)|} \cdot (1 - p_n)^{n - |R(\mcA)|},
\]
so, informally speaking, $\mbbP_n^\mbbG$ is the probability distribution on $\mbW_n$ such that, for all $i \in [n]$,
the probability that $R(i)$ holds is $n^{-\alpha}$, independently of whether $R(j)$ holds for $j \neq i$.
For {\em every} quantifier-free MLN $\mbbM$ over $\sigma$ there is $n_\mbbM \in \mbbN$
such that for all integers $n \geq n_\mbbM$,
$\mbbP_n^\mbbM \neq \mbbP_n^\mbbG$.
\end{theor}

\begin{rem}\label{remark on MLNs with and without quantifiers}{\rm
One may ask if, for every MLN $\mbbM$ over $\sigma$ there is a {\em quantifier-free} MLN $\mbbM'$ over $\sigma$
such that for all sufficiently large $n$, $\mbbP_n^\mbbM = \mbbP_n^{\mbbM'}$, or alternatively
$\mbbP_n^\mbbM \approx \mbbP_n^{\mbbM'}$ with arbitrarily small error for all large enough $n$.
I conjecture that the answer is negative.
To give an idea of why, let $\psi(x, y)$ be the formula
\begin{align*}
&x=x \wedge y=y \ \wedge  \\
&\exists u, v, w \big(R(u) \wedge R(v) \wedge R(w) \wedge u \neq v \wedge u \neq w \wedge v \neq w 
\ \wedge \\
&\forall z \big(R(z) \rightarrow (z = u \vee z = v \vee z = w)\big)\big).
\end{align*}
Then let $\mbbM_3 := \{(\psi(x, y), w)\}$ where $w > 0$.
It follows that if $\mcA \in \mbW_n^3$ then $\mu_n^{\mbbM_3}(\mcA) = 2^{wn^2}$,
and if $\mcA \in \mbW_n \setminus \mbW_n^3$ then $\mu_n^{\mbbM_3}(\mcA) = 1$.
Note also that $|\mbW_n| = 2^n$.
By straightforward arguments it follows that
$\lim_{n\to\infty} \mbbP_n^{\mbbM_3}\big(\mbW_n^3\big) = 1$.
Suppose that $\mbbM$ is a quantifier-free MLN over $\sigma$.
Due to the character of the function 
(in the proof of Theorem~\ref{characterization of quantifier-free MLNs for colourings})
that determines $\mbbP_n^\mbbM\big(\mbW_n^3\big)$
I do not believe that $\mbbP_n^\mbbM$ can concentrate almost all of its probability mass on $\mbW_n^3$ as $n\to\infty$. 
But to verify (or refute) this conjecture seems to require a detailed technical analysis which I leave for another occasion.
}\end{rem}

\subsection{MLNs versus LBNs over richer languages}\label{MLNs versus LBNs over richer languages}

\noindent
In Section~\ref{Quantifier free MLNs over a signature with only one relation symbol}
we considered a language with only one relation symbol which was unary.
Here we give a result in the same spirit as 
Theorems~\ref{MLN with two convergence points} and~\ref{limits of MLNs}
which tell that LBNs and quantifier-free MLNs are incomparable 
over any language which has at least one relation symbol of arity~1.

\begin{theor}\label{MLN versus LBN generally}
Let $\tau$ be a language with at least one relation symbol of arity 1.\\
(a) There is a quantifier-free MLN $\mbbM$ over $\tau$ such that, for every LBN $\mbbG$ over $\tau$,
if $n \in \mbbN$ is sufficiently large then $\mbbP_n^\mbbM \neq \mbbP_n^\mbbG$.\\
(b) There is an LBN $\mbbG$ over $\tau$ such that for every quantifier-free MLN $\mbbM$ over $\tau$,
if $n \in \mbbN$ is sufficiently large then $\mbbP_n^\mbbG \neq \mbbP_n^\mbbM$.
\end{theor}

\noindent
{\bf Proof.}
Let $R \in \tau$ have arity 1. 
If $R$ is the only symbol of $\tau$ then the result follows from 
Theorems~\ref{MLN with two convergence points}
and~\ref{limits of MLNs}.
So now suppose that $\tau = \{R\} \cup \sigma'$ where $\sigma'$ is nonempty.
Let $\sigma = \{R\}$.
Let $\mbW_n(\sigma)$, $\mbW_n(\tau)$, and $\mbW_n(\sigma')$ denote the
set of $\sigma$-structures, $\tau$-structures, respectively $\sigma'$-structures with domain $[n]$
(so $\mbW_n(\tau) = \mbW_n$).

(a) Let $\varphi_0(x, y), \varphi_1(x, y)$ and $\varphi_2(x, y)$ be the formulas
from Theorem~\ref{MLN with two convergence points},
and let $w_0 = w_2 = 1$ and $w_1 = 0$ just as in the same theorem.
Let 
\[
\mbbM_\sigma = \{(\varphi_0(x, y), w_0), (\varphi_1(x, y), w_1), (\varphi_2(x, y), w_2)\}
\]
where {\em we consider $\mbbM_\sigma$ as an MLN over $\sigma$}
(so $\mbbM_\sigma$ is the same MLN as in Theorem~\ref{MLN with two convergence points}).
Then let 
\[
\mbbM_\tau = \{(\varphi_0(x, y), w_0), (\varphi_1(x, y), w_1), (\varphi_2(x, y), w_2)\}
\]
where {\em we consider $\mbbM_\tau$ as an MLN over $\tau$}.
So for every $n$, $\mbbP_n^{\mbbM_\sigma}$ is a probability distribution on $\mbW_n(\sigma)$
and $\mbbP_n^{\mbbM_\tau}$ is a probability distribution on $\mbW_n(\tau)$.

For each $n$ and $\mcB \in \mbW_n(\sigma)$ define
\[
\mbX_n(\mcB) = \{ \mcA \in \mbW_n(\tau) : \mcA \uhrc \sigma = \mcB \}.
\]
For each $n$ and $\mcB \in \mbW_n(\sigma')$ define
\[
\mbY_n(\mcB) = \{ \mcA \in \mbW_n(\tau) : \mcA \uhrc \sigma' = \mcB \}.
\]
For all $\mcA \in \mbW_n(\tau)$ we have
\[
\mu_n^{\mbbM_\tau}(\mcA) = \mu_n^{\mbbM_\sigma}(\mcA \uhrc \sigma) \ \ 
\text{ and hence } \ \ 
\mu_n^{\mbbM_\tau}(\mbW_n(\tau)) = \mu_n^{\mbbM_\sigma}(\mbW_n(\sigma)) \cdot |\mbW_n(\sigma')|.
\]
It follows that for all $\mcA \in \mbW_n(\tau)$,
\begin{equation}\label{characterization of mu(A)-tau}
\mbbP_n^{\mbbM_\tau}(\mcA) = 
\frac{\mu_n^{\mbbM_\tau}(\mcA)}{\mu_n^{\mbbM_\tau}(\mbW_n(\tau))} = 
\frac{\mu_n^{\mbbM_\sigma}(\mcA \uhrc \sigma)}{\mu_n^{\mbbM_\sigma}(\mbW_n(\sigma)) \cdot |\mbW_n(\sigma')|}
= \frac{\mbbP_n^{\mbbM_\sigma}(\mcA \uhrc \sigma)}{|\mbW_n(\sigma')|}.
\end{equation}
It follows from~(\ref{characterization of mu(A)-tau}) that
\begin{align}\label{probability of particular interpretation of R}
&\text{for all } \mcB \in \mbW_n(\sigma),  \ 
\mbbP_n^{\mbbM_\tau}\big(\mbX_n(\mcB)\big) 
= \sum_{\mcA \in \mbX_n(\mcB)} \mbbP_n^{\mbbM_\tau}(\mcA) \\
&= \sum_{\mcA \in \mbX_n(\mcB)} \frac{\mbbP_n^{\mbbM_\sigma}(\mcB)}{|\mbW_n(\sigma')|}
= \mbbP_n^{\mbbM_\sigma}(\mcB) \qquad \text{ because } |\mbX_n(\mcB)| = |\mbW_n(\sigma')|.
\nonumber
\end{align}
By using~(\ref{characterization of mu(A)-tau}) again we get
\begin{align}\label{probability of particular interpretation of sigma'}
&\text{for all } \mcB \in \mbW_n(\sigma'),  \ 
\mbbP_n^{\mbbM_\tau}\big(\mbY_n(\mcB)\big) 
= \sum_{\mcA \in \mbY_n(\mcB)} \mbbP_n^{\mbbM_\tau}(\mcA) \\
&= \sum_{\mcA \in \mbY_n(\mcB)} \frac{\mbbP_n^{\mbbM_\sigma}(\mcA \uhrc \sigma)}{|\mbW_n(\sigma')|} 
= \frac{ \underset{\mcA \in \mbW_n(\sigma)}{\sum} \mbbP_n^{\mbbM_\sigma}(\mcA)}{|\mbW_n(\sigma')|} 
= \frac{1}{\big|\mbW_n(\sigma')\big|}.
\nonumber
\end{align}
From~(\ref{characterization of mu(A)-tau}), (\ref{characterization of mu(A)-tau}) 
and~(\ref{probability of particular interpretation of sigma'}) 
it follows that
\begin{align}\label{independence between R and sigma'}
&\text{for all } \mcA\in \mbW_n(\tau),  \ 
\mbbP_n^{\mbbM_\tau}(\mcA) = 
\mbbP_n^{\mbbM_\sigma}(\mcA \uhrc \sigma) \cdot \frac{1}{\big|\mbW_n(\sigma')\big|} \\
&= \mbbP_n^{\mbbM_\tau}\big(\mbX_n(\mcA \uhrc \sigma)\big) \cdot 
\mbbP_n^{\mbbM_\tau}\big(\mbY_n(\mcA \uhrc \sigma')\big).
\nonumber
\end{align}
This means that the interpretation of $R$ is independent (with respect to $\mbbP_n^\mbbM$) 
of the interpretation of other relation symbols.

Towards a contradiction, suppose that $\mbbG_\tau$ is an LBN over $\tau$ and that
$\mbbP_n^{\mbbG_\tau} = \mbbP_n^{\mbbM_\tau}$ for infinitely many $n$.
By Lemma~\ref{removing redundant edges}
we may, without loss of generality, assume that $\mbbG_\tau$ has no asymptotically redundant edge.
For the remainder of the argument suppose that $n$ is such that 
$\mbbP_n^{\mbbG_\tau} = \mbbP_n^{\mbbM_\tau}$.

Then, by~(\ref{probability of particular interpretation of R}),
(\ref{probability of particular interpretation of sigma'})
and~(\ref{independence between R and sigma'})
we get
\begin{align}\label{transfer to G-tau}
&\text{for all } \mcB \in \mbW_n(\sigma),  \ 
\mbbP_n^{\mbbG_\tau}\big(\mbX_n(\mcB)\big) = 
\mbbP_n^{\mbbM_\tau}\big(\mbX_n(\mcB)\big) =
\mbbP_n^{\mbbM_\sigma}(\mcB), \\
&\text{for all } \mcB \in \mbW_n(\sigma'),  \ 
\mbbP_n^{\mbbG_\tau}\big(\mbY_n(\mcB)\big) = \mbbP_n^{\mbbM_\tau}\big(\mbY_n(\mcB)\big) = 
\frac{1}{\big|\mbW_n(\sigma')\big|}, \text{ and} \nonumber\\
&\text{for all } \mcA\in \mbW_n(\tau),  \ 
\mbbP_n^{\mbbG_\tau}(\mcA) = 
\mbbP_n^{\mbbM_\tau}(\mcA) = 
 \mbbP_n^{\mbbM_\tau}\big(\mbX_n(\mcA \uhrc \sigma)\big) \cdot 
\mbbP_n^{\mbbM_\tau}\big(\mbY_n(\mcA \uhrc \sigma')\big) 
\nonumber \\
&=  \mbbP_n^{\mbbG_\tau}\big(\mbX_n(\mcA \uhrc \sigma)\big) \cdot 
\mbbP_n^{\mbbG_\tau}\big(\mbY_n(\mcA \uhrc \sigma')\big).
\nonumber
\end{align}
The above means that, with respect to $\mbbP_n^{\mbbG_\tau}$, 
the interpretation of $R$ is independent from the interpretations of
the other relation symbols.
Since we assume that $\mbbG_\tau$ has no asymptotically redundant edge 
and $n$ can be as large as we like it follows that $R$ has no 
parent in $\mbbG_\tau$.
Let $S_R$ be the map associated to $R$ by $\mbbG_\tau$ according to
Definition~\ref{definition of LBN} of an LBN.
Then $S_R$ is a map from pairs $(\mcA, a)$ to $[0, 1]$, where $\mcA$ is a finite
$\es$-structure, that is, a finite set, and $a$ is a member of that set.
Now construct an LBN $\mbbG_\sigma$ over $\sigma = \{R\}$ 
by letting the same $S_R$ be the map that $\mbbG_\sigma$
associates to $R$.
Then it is clear from Definition~\ref{semantics of LBN}
and~(\ref{transfer to G-tau})
that, for all $\mcA \in \mbW_n(\sigma)$,
\[
\mbbP_n^{\mbbG_\sigma}(\mcA) = \mbbP_n^{\mbbG_\tau}\big(\mbX_n(\mcA)\big) 
= \mbbP_n^{\mbbM_\tau}\big(\mbX_n(\mcA)\big) = \mbbP_n^{\mbbM_\sigma}(\mcA),
\]
so $\mbbP_n^{\mbbG_\sigma} =  \mbbP_n^{\mbbM_\sigma}$.
As this holds for all sufficiently large $n$ we have a contradiction to
Theorem~\ref{MLN with two convergence points}.

(b) Let $\alpha \in (0, 1)$ and $p_n := n^{-\alpha}$ for all $n \in \mbbN^+$.
 Let $\mbbG_\tau$ be the LBN over $\tau$ which is defined as follows.
The DAG of $\mbbG_\tau$ has no edge, so $\mr{par}_{\mbbG_\tau}(Q) = \es$ for all $Q \in \tau$.
For every finite $\es$-structure $\mcA$ and $a \in A$, $S_R(\mcA, a) = p_{|A|}$.
For every $Q \in \sigma'$, every finite $\es$-structure $\mcA$ and every
$\bar{a} \in A^{\mr{ar}(Q)}$, $S_Q(\mcA, \bar{a}) = 1/2$.
It follows that for all $n$ and $\mcA \in \mbW_n(\tau) $,
\[
\mbbP_n^{\mbbG_\tau}(\mcA) = 
p_n^{|R(\mcA)|} \cdot (1 - p_n)^{n - |R(\mcA)|}
\cdot \frac{1}{|\mbW_n(\sigma')|}.
\]
Let $\mbbG_\sigma$ be the LBN over $\sigma$ defined by letting the map associated to $R$ be
$S_R(\mcA, a)$, the same map as $\mbbG_\tau$ associates to $R$.
Hence 
$\mbbP_n^{\mbbG_\sigma}(\mcA) = 
p_n^{|R(\mcA)|} \cdot (1 - p_n)^{n - |R(\mcA)|}$
for all $\mcA \in \mbW_n(\sigma)$.

For $\mcB \in \mbW_n(\sigma')$ let $\mbY_n(\mcB)$ be defined as in part~(a).
For each $n \in \mbbN^+$ let $\mcB_n \in \mbW_n(\sigma')$ be such that for all
$Q \in \sigma'$, $Q^{\mcB_n} = \es$, so $\mcB_n \models \neg Q(\bar{a})$
for all $\bar{a} \in [n]^{\mr{ar}(Q)}$.
It follows that for all $\mcA \in \mbY_n(\mcB_n)$
\begin{equation}\label{connection between G-tau and G-sigma}
\mbbP_n^{\mbbG_\tau}\big(\mcA \ \big| \ \mbY_n(\mcB_n)\big) = 
\mbbP_n^{\mbbG_\sigma}\big(\mcA \uhrc \sigma\big) = 
p_n^{|R(\mcA)|} \cdot (1 - p_n)^{n - |R(\mcA)|}.
\end{equation}

Suppose that $\mbbM_\tau$ is a quantifier-free MLN over $\tau$.
By Lemma~\ref{normalized quantifier-free MLN},
we may, without loss of generality, assume that every soft constraint of $\mbbM_\tau$ is
a maximal consistent conjunction of $\tau$-literals.
Let $\varphi_i(x_1, \ldots, x_{k_i})$, $i = 1, \ldots, m$, be an enumeration of all
soft constraints $\varphi(x_1, \ldots, x_k)$ of $\mbbM_\tau$ such that 
$\neg Q(x_{j_1}, \ldots, x_{j_{\mr{ar}(Q)}})$ is a conjunct of $\varphi(x_1, \ldots, x_k)$
for every $Q \in \sigma'$ and every choice of $x_{j_1}, \ldots, x_{j_{\mr{ar}(Q)}} \in \{x_1, \ldots, x_k\}$.
Each $\varphi_i(x_1, \ldots, x_{k_i})$ can be written on the form
\[
\varphi_{i, 1}(x_1, \ldots, x_{k_i}) \wedge \varphi_{i, 2}(x_1, \ldots, x_{k_i})
\]
where $\varphi_{i, 1}$ is a maximal consistent conjunction of $\sigma$-literals and
$\varphi_{i, 2}$ is a maximal consistent conjunction of $\sigma'$-literals.
Let $w_i$ be the weight associated to $\varphi_i$ by $\mbbM_\tau$.
Define
\[
\mbbM_\sigma = \big\{(\varphi_{i, 1}(x_1, \ldots, x_{k_i}), w_i) : i = 1, \ldots, m\big\}.
\]
Then, for all $\mcA \in \mbY_n(\mcB_n)$,
\begin{equation}\label{connection between M-tau and M-sigma}
\mu_n^{\mbbM_\tau}(\mcA) = \mu_n^{\mbbM_\sigma}(\mcA \uhrc \sigma) \ \text{ so } \ 
\mbbP_n^{\mbbM_\tau}\big(\mcA \ \big| \ \mbY_n(\mcB_n)\big) = 
\mbbP_n^{\mbbM_\sigma}(\mcA \uhrc \sigma).
\end{equation}
Suppose that $\mbbP_n^{\mbbM_\tau} = \mbbP_n^{\mbbG_\tau}$.
It follows from
(\ref{connection between G-tau and G-sigma}) and~(\ref{connection between M-tau and M-sigma})
 that for all $\mcA \in \mbY_n(\mcB_n)$,
\[
\mbbP_n^{\mbbM_\sigma}(\mcA \uhrc \sigma) = 
\mbbP_n^{\mbbM_\tau}\big(\mcA \ \big| \ \mbY_n(\mcB_n)\big) = 
\mbbP_n^{\mbbG_\tau}\big(\mcA \ \big| \ \mbY_n(\mcB_n)\big) = 
\mbbP_n^{\mbbG_\sigma}(\mcA \uhrc \sigma).
\]
Since every $\mcA \in \mbW_n(\sigma)$ has an expansion which belongs to $\mbY_n(\mcB_n)$
it follows that for all $\mcA \in \mbW_n(\sigma)$,
\[
\mbbP_n^{\mbbM_\sigma}(\mcA) = \mbbP_n^{\mbbG_\sigma}(\mcA)
= p_n^{|R(\mcA)|} \cdot (1 - p_n)^{n - |R(\mcA)|}.
\]
If this holds for arbitrarily large $n$ we have a contradiction to
Theorem~\ref{limits of MLNs}.
Therefore there cannot be infinitely many $n$ such that $\mbbP_n^{\mbbM_\tau} = \mbbP_n^{\mbbG_\tau}$.
\hfill $\square$

\section{Weight versus domain size in the limit}\label{graphs with maximum degree}

\noindent
We recall that weights of soft constraints range over the non-negative reals, and domain sizes range over the
positive integers. Given an MLN $\mbbM_w = \{(\varphi(\bar{x}), w)\}$ we get a sequence of probability distributions
$(\mbbP_n^{\mbbM_w} : n \in \mbbN^+)$ representing the ``domain size dimension''.
But we can also consider the ``weight dimension'' $(\mbbP_n^{\mbbM_w} : w \in \mbbR^{\geq 0})$ where we let
$w$ vary and $n$ stay fixed.
One may ask if these two dimensions are somehow correlated. 
For example, does a large enough weight $w$ guarantee that there is $c > 0$ such that for all large enough $n$,
$\mbbP_n^{\mbbM_w}(\forall \bar{x} \varphi(\bar{x})) := 
\mbbP_n^{\mbbM_w}\big(\{ \mcA \in \mbW_n : \mcA \models \forall \bar{x} \varphi(\bar{x})\}\big) \geq c$?
We will give a negative answer to the question which shows that a MLNs may behave as differently as 
possible along the
two dimensions. In particular, the choice of weight $w$ may have no impact at all
on $\lim_{n\to\infty} \mbbP_n^{\mbbM_w}(\forall \bar{x} \varphi(\bar{x}))$.

In this section we let $\sigma = \{R\}$ where $R$ is a relation symbol of arity 2
which is always interpreted
as an irreflexive and symmetric relation,
so $\mbW_n$ is the set of all undirected graphs without loops, from now on simply called graphs, 
with vertex set $[n] := \{1, \ldots, n\}$.

{\em For the rest of the section fix some $\Delta \in \mbbN^+$.}
Let $\varphi(x_1, \ldots, x_{\Delta + 2})$ be the formula
\[
\bigvee_{2 \leq i < j \leq \Delta + 2} x_i = x_j \ \vee \ \bigvee_{i = 2}^{\Delta + 2} \neg R(x_1, x_i),
\]
so $\varphi(x_1, \ldots, x_{\Delta + 2})$ expresses that if all $x_2, \ldots, x_{\Delta + 2}$ are different then
$x_1$ is {\em not} adjacent to all $x_2, \ldots, x_{\Delta + 2}$.
It follows that if $\mcA \in \mbW_n$ and $\mcA \models \varphi(a_1, \ldots, a_{\Delta + 2})$ for all choices
of $a_1, \ldots, a_{\Delta + 2} \in [n]$, then every vertex of $\mcA$ has degree at most $\Delta$ and in this
case we say that $\mcA$ {\em has maximum degree at most $\Delta$}.
Let 
\[
\mbbM_w := \big\{(\varphi(x_1, \ldots, x_{\Delta + 2}), w)\big\} \quad \text{ where } w \geq 0.
\]
If $w > 0$ we can think of $\mbbM_w$ as modelling a situation where it is less likely that someone
has more than $\Delta$ friends than that the person has at most $\Delta$ friends;
or alternatively, less likely that an atom binds to more than $\Delta$ atoms than to at most
$\Delta$ atoms.

For all $n \in \mbbN^+$ let
\begin{align*}
\mbOm_n^\Delta &:= 
\{ \mcA \in \mbW_n : \mcA \models \forall x_1, \ldots, x_{\Delta + 2} \varphi(x_1, \ldots, x_{\Delta + 2})\} \\
&= \big\{ \mcA \in \mbW_n : \text{ $\mcA$ has maximum degree at most $\Delta$} \big\}.
\end{align*}

\noindent
A direct application of \cite[Proposition~4.3]{RD} to $\mbbM_w$ gives the following:

\begin{theor}\label{limit as w tends to infinity}
For every choice of $n \in \mbbN^+$, 
$\lim_{w\to\infty} \mbbP_n^{\mbbM_w}\big(\mbOm_n^\Delta\big) = 1$.
\end{theor}

\noindent
In Appendix~\ref{proof of probability of max degree Delta tends to 0}, we prove the
following:

\begin{theor}\label{probability of max degree Delta tends to 0}
For every choice of the weight $w \geq 0$, 
$\lim_{n\to\infty} \mbbP_n^{\mbbM_w}\big(\mbOm_n^{\Delta}\big) = 0$.
\end{theor}

\begin{rem}\label{remark on knowledge transfer for maximum degree}{\rm 
(Implications for knowledge transfer) 
Let $\varepsilon > 0$ be small.
Suppose that, for some (fixed) $m$, we have calibrated the weight $w$ so that 
$\mbbP_m^{\mbbM_w}\big(\mbOm_m^\Delta\big) \geq 1 - \varepsilon$.
By~\cite[Proposition~4.3]{RD}, this is possible by choosing $w$ large enough (since $m$ is fixed).
The justification for this choice of $w$ may be that the weight $w$ was learned from
examples with domain size $m$ such that with high probability an example satisfied
$\forall x_1, \ldots,  x_{\Delta + 2} \varphi(x_1, \ldots, x_{\Delta + 2})$.
Suppose that we now want to use $\mbbM_w$ to make inferences on $\mbW_n$
with $\mbbP_n^{\mbbM_w}$ for some $n$
which is much larger than $m$.
Theorem~\ref{probability of max degree Delta tends to 0}
implies that if $n$ is large enough then 
$\mbbP_n^{\mbbM_w}\big(\mbOm_n^\Delta\big) \leq \varepsilon$.
As $\mbbP_m^{\mbbM_w}\big(\mbOm_m^\Delta\big) \geq 1 - \varepsilon$
we can {\em not} extrapolate knowledge about the probability of the event that no vertex has more
than $\Delta$ neighbours from a domain with size $m$ to a domain with size $n$.
}\end{rem}

\begin{rem}\label{does not help to add quantifiers} {\rm
We do not avoid getting the limit 0 in
Theorem~\ref{probability of max degree Delta tends to 0}
by reducing the number of free variables by the use of quantifiers.
For example, if we replace $\varphi(x_1, \ldots, x_{\Delta + 2})$ in $\mbbM_w$ by, say, the soft constraint
\[
\psi(x_1) := 
\forall x_2, \ldots, x_{\Delta + 2} 
\Big(\bigvee_{2 \leq i < j \leq \Delta + 2} x_i = x_j \ \vee \ \bigvee_{i = 2}^{\Delta + 2} \neg R(x_1, x_i)\Big)
\]
then we can argue essentially as in the proof of
Theorem~\ref{probability of max degree Delta tends to 0}
(in Appendix~\ref{proof of probability of max degree Delta tends to 0}), 
by just replacing `$n^{\Delta + 2}$' in that proof by `$n$'  and letting $\tau := \Delta + 3$ in the same proof.

The reason that the limit is 0 for every choice of $w$ seems to be that, although, for some $d > c > 0$, 
there are between $2^{cn}$ and $2^{dn}$ graphs
with vertex set $[n]$ and with maximum degree $\Delta$, 
we also have $2^{dn} \big/ |\mbW_n| = 2^{dn} \big/ 2^{\frac{n^2 - n}{2}} \to 0$ (exponentially fast) as $n \to \infty$.
So even if each graph with maximum degree at most $\Delta$ gets much higher weight than a graph that does not satisfy this
condition, there may simply be too few graphs of the first sort to make its total weight comparable to that of  the second sort.
Another reason could be that even if $w$ is very large, there may be (too) many graphs with maximum degree larger than $\Delta$,
but with only few failures of the soft constraint and then the weight of each such graph need not be much lower 
(in proportional terms)
than the weight of a graph with maximum degree $\Delta$.

I think that the proof of Theorem~\ref{probability of max degree Delta tends to 0}
can be modified to show that, for every fixed $k$, the probability that there are at most $k$ $(\Delta +2)$-tuples that
violate the soft constraint $\varphi(x_1, \ldots, x_{\Delta + 2})$ tends to 0 as $n \to \infty$ (but the combinatorics would
become messier).
}\end{rem}

\section{Conclusion}

\noindent
We have considered Markov logic networks (MLNs) $\mbbM$ and the probability distribution $\mbbP_n^\mbbM$
on the set $\mbW_n$ of structures, or ``possible worlds'', with domain $[n] := \{1, \ldots, n\}$ 
where $\mbbP_n^\mbbM$ is determined by $\mbbM$.
The problem of how  $\mbbP_n^\mbbM$ behaves, as $n \to \infty$, has been addressed.
Only few and rather weak results on the domain size asymptotics of MLNs exist
\cite{JBB, Mittal, Poole, Wei25}
and this topic is of importance for understanding knowledge transfer and scalability of learning and inference with MLNs.

We have seen that with mild assumptions on an MLN $\mbbM$ with one soft constraint with an
arbitrary positive weight the distribution $\mbbP_n^\mbbM$ will
behave quite differently from the uniform distribution $\mbbP_n^{uni}$ on $\mbW_n$ for all large $n$.

For a language $\sigma = \{R\}$ where $R$ has arity 1, we have described the asymptotic behaviour of
$\mbbP_n^\mbbM$, as $n \to \infty$, when $\mbbP_n^\mbbM$ is determined by a quantifier-free 
MLN $\mbbM$ over $\sigma$.
It shows that the weights of the soft constraints of $\mbbM$ do influence, in fact determine, 
the limit behaviour of $\mbbP_n^\mbbM$ as $n \to \infty$.
We have also proved that if the language $\sigma$ contains at least one
relation symbol of arity 1, then quantifier-free MLNs and Lifted Bayesian networks (LBNs) over $\sigma$ have
incomparable expressive strength.
That is, there is a quantifier-free MLN $\mbbM$ over $\sigma$  such that 
for every LBN $\mbbG$ over $\sigma$,
$\mbbM$ and $\mbbG$ determine different distributions on $\mbW_n$ for all sufficiently large $n$;
and conversely, there is an LBN $\mbbG$ over $\sigma$  such that 
for every quantifier-free MLN $\mbbM$ over $\sigma$,
$\mbbG$ and $\mbbM$ determine different distributions on $\mbW_n$ for all sufficiently large $n$.

It is known from \cite[Proposition 4.3]{RD} that if $\mbbM_w := \{(\varphi(x_1, \ldots, x_k), w)\}$
is an MLN and $\varphi$ is quantifier-free and consistent then, for every fixed $n$,
we have $\lim_{w\to\infty}$ $\mbbP_n^{\mbbM_w}\big(\forall x_1, \ldots, x_k$ $\varphi(x_1, \ldots, x_k)\big) = 1$
(i.e. the probability that all $k$-tuples satisfy $\varphi(x_1, \ldots, x_k)$ tends to 1 as the weight $w \to \infty$).
We have proved that if $k \geq 3$ and $\varphi(x_1, \ldots, x_k)$
expresses that $x_1$ is not connected to all of $x_2, \ldots, x_k$, then for every fixed $w \geq 0$,
$\lim_{n\to\infty}$ $\mbbP_n^{\mbbM_w}\big(\forall x_1, \ldots, x_k$ $\varphi(x_1, \ldots, x_k)\big) = 0$.
So in the limit, the weight dimension and the domain size dimension may behave completely differently.

The present and earlier studies show that the domain size asymptotics of an MLN depends heavily on
which soft constraints the MLN uses. To scale the weights of soft constraints
by a number determined by the domain size does not in general change this picture
(recall the discussion in Section~\ref{Related work}).
At the current state of affairs a general theory of domain size asymptotics of MLNs
is indeed far away and the next realistic step appears to be to study well chosen cases of 
more or less concrete MLNs. It is natural to start with simple languages and simple soft constraints
and, after understanding these sufficiently well, proceed to more complex languages and 
more complex soft constraints.
So in \cite{TK1} Tousinejad and the author give a quite detailed description of the domain size asymptotics of
MLNs over a language with only two relation symbols, one of arity 1 and the other of arity 2, and
where the soft constraints are any quantifier-free formulas with 2 variables (such MLNs can be 
transformed to the specific kind considered in \cite{TK1} without changing the distribution);
a striking feature is that with the scaled MLN semantics the description is much simpler than with
the standard unscaled semantics.
I conjecture that a similar analysis (as in \cite{TK1}),  but more complex, holds for any language with only
relation symbols of arity at most 2, and quantifier-free soft constraints with 2 variables.
It is more challenging, even under the assumption that all relations have arity at most 2, 
to understand the domain size asymptotics of MLNs with quantifier-free soft constraints that have more variables than
the maximal arity of the relations in the language. The study \cite{TK2} is a step in this direction.
But beyond it the area is wide open for research.
What soft constraints should be considered first? 
Perhaps such which are not too complicated
and which have an intuitive meaning in some branch of science, technology or the social sphere?

\subsection*{Acknowledgements}
I thank two anonymous reviewers for remarks that improved the exposition of this article,
Yasmin Tousinejad for finding some errors in previous versions of this article, and
Prof. Stephan Wagner for helping me out and proving 
Proposition~\ref{when the weighted binomial distribution is constant}.

For this version of the article, unlike two earlier versions 
(posted on \url{https://arxiv.org/abs/2509.04192}), 
generative AI was used to check correctness of the results
and it did also provide shorter and more elegant proofs in some cases. 
In this version, part of the proof of Theorem~\ref{MLN with two convergence points},
part 2(b) of the proof of Theorem~\ref{characterization of quantifier-free MLNs for colourings},
and the whole proof of Theorem~\ref{limits of MLNs} was found by generative AI.
However, the presented proofs are not copied literally from the AI's suggestions but are reformulated by the 
author to make them easier to follow.

This study was partially supported by the Swedish Research Council, grant 2023-05238\_VR.

\appendix

\section{Auxiliary results}

\noindent
In this appendix some auxiliary results are stated which will be used in the proofs of the main results.
The following result about independent Bernoulli trials 
is a direct consequence of \cite[Corollary~A.1.14]{AS} which in turn follows from a bound given by
Chernoff \cite{Che}:

\begin{lem}\label{independent bernoulli trials}
Let $Z$ be the sum of $n$ independent 0/1-valued random variables, each one with probability $p$ of having the value 1,
where $p > 0$.
For every $\varepsilon > 0$ there is $c_\varepsilon > 0$, depending only on $\varepsilon$, such that the probability that
$|Z - pn| > \varepsilon p n$ is less than $2 e^{-c_\varepsilon p n}$.
\end{lem}

\begin{prop}\label{when the weighted binomial distribution is constant} {\rm (due to Prof. Stephan Wagner, private communication)} \\
Let $w_0, \ldots w_k \in \mbbR$.
The polynomial
\[
f(x) = \sum_{s=0}^k w_s \binom{k}{s} x^s(1-x)^{k-s}.
\]
is constant if and only if $w_0 = \ldots = w_k$.
\end{prop}

\noindent
{\bf Proof.}
If the weights are all equal, i.e., $w_0 = w_1 = \cdots = w_k = w$, then the polynomial simplifies to
\begin{equation*}
f(x) = w \sum_{s=0}^k \binom{k}{s} x^s (1-x)^{k-s} = w (x + 1-x)^k = w.
\end{equation*}
So we only need to prove the converse. Suppose that the polynomial is constant. We first rewrite it by grouping terms according to the exponent of $x$:
\begin{align*}
f(x) &= \sum_{s=0}^k w_s \binom{k}{s} x^s (1-x)^{k-s} = \sum_{s=0}^k w_s \binom{k}{s} x^s \sum_{r=0}^{k-s} \binom{k-s}{r} (-x)^r \\
&= \sum_{s=0}^k w_s \sum_{r=0}^{k-s}  \binom{k}{s} \binom{k-s}{r} (-1)^r x^{r+s} = \sum_{s=0}^k w_s \sum_{r=0}^{k-s}  \frac{k!(k-s)!}{s!(k-s)!r!(k-s-r)!} (-1)^r x^{r+s} \\
&= \sum_{s=0}^k \sum_{r=0}^{k-s}  \binom{k}{r+s} \binom{r+s}{s} (-1)^r w_s x^{r+s} = \sum_{n=0}^{k} \sum_{s=0}^n \binom{k}{n} \binom{n}{s} (-1)^{n-s} w_s x^n \\
&= \sum_{n=0}^{k} \binom{k}{n} \Big( \sum_{s=0}^n \binom{n}{s} (-1)^{n-s} w_s \Big) x^n.
\end{align*}
So for the polynomial to be constant, we must have
\begin{equation*}
\sum_{s=0}^n \binom{n}{s} (-1)^{n-s} w_s = 0
\end{equation*}
for all $n > 0$. Now we show by strong induction on $n$ that this implies $w_n = w_0$ for all $n \geq 0$. This is trivial for $n = 0$, so we proceed with the induction step. For $n > 0$, the induction hypothesis gives us
\begin{align*}
0 &= \sum_{s=0}^n \binom{n}{s} (-1)^{n-s} w_s = w_n + \sum_{s=0}^{n-1} \binom{n}{s} (-1)^{n-s} w_0 \\
&= w_n + w_0 \Big(-1 + \sum_{s=0}^{n} \binom{n}{s} (-1)^{n-s} \Big) = w_n + w_0 \Big( -1 + (1-1)^n \Big) \\
&= w_n - w_0.
\end{align*}
Thus $w_n = w_0$, completing the induction.
\hfill $\square$

\section{Proof of Theorem~\ref{MLN with two convergence points}}\label{proof of MLN with two convergence points}

\noindent
Let $\sigma = \{R\}$ where $R$ is a relation symbol of arity 1 and let $\mbbM$ be the MLN
that is defined in Theorem~\ref{MLN with two convergence points}.
Recall the notation
\[
\mbW_n^m := \big\{\mcA \in \mbW_n : |R(\mcA)| = m\}.
\]
In this proof, let us simplify notation by letting
$\mu_n := \mu_n^\mbbM$ and $\mbbP_n := \mbbP_n^\mbbM$ for all $n$,
(where $\mu_n^\mbbM$ and $\mbbP_n^\mbbM$ are as in 
Definition~\ref{definition of distribution defined by an MLN}).
Everywhere in the proof we assume that $n \geq 2$.

Suppose that $\mcA \in \mbW_n$ and $|R(\mcA)| = m$, so $\mcA$ has exactly $m$ coloured elements.
The number of ordered pairs of (not necessarily distinct) elements from $[n]$ (the domain of $\mcA$)
such that 
\begin{itemize}
\item both entries are coloured is $m^2$,
\item exactly one entry is coloured is $2m(n-m)$, and
\item both entries are uncoloured is $(n-m)^2$.
\end{itemize}
It follows from the definition of $\mbbM$ in
Theorem~\ref{MLN with two convergence points},
which sets $w_0 := w_2 := 1$ and $w_1 := 0$,
and from
Definition~\ref{definition of distribution defined by an MLN} of $\mu_n$ that 
\begin{align*}
\mu_n(\mcA) &= \exp_2\big(w_0 m^2 + 2w_1m(n-m) + w_2(n-m)^2\big) 
= \exp_2\big(m^2 + (n-m)^2\big) \\
&= \exp_2\big(n^2 - 2m(n-m)\big).
\end{align*}
Hence 
\[
\mu_n\big(\mbW_n^m\big) = \binom{n}{m} \exp_2\big(n^2 - 2m(n-m)\big).
\]
If $1 \leq m \leq n-1$ then
\[
m(n-m) - (n-1) = (m-1)(n-m-1) \geq 0
\]
so 
\[
m(n-m) \geq n-1 \quad \text{ provided that $1 \leq m \leq n-1$.}
\]
We have $\mu_n\big(\mbW_n^0\big) = 2^{n^2} = \mu_n\big(\mbW_n^n\big)$, 
so  $\mu_n\big(\mbW_n^0 \cup \mbW_n^n\big) = 2 \cdot 2^{n^2}$ and we get
\begin{align*}
&\mbbP_n\bigg(\bigcup_{m=1}^{n-1} \mbW_n^m\bigg) = 
\frac{\mu_n\Big(\bigcup_{m=1}^{n-1} \mbW_n^m\Big)}{\mu_n\big(\mbW_n\big)} 
\leq \frac{\mu_n\Big(\bigcup_{m=1}^{n-1} \mbW_n^m\Big)}{\mu_n\big(\mbW_n^0 \cup \mbW_n^n\big)} \\
&\leq \frac{\sum_{m=1}^{n-1} \binom{n}{m} 2^{n^2 - 2m(n-m)}}{2 \cdot 2^{n^2}} =
\frac{1}{2} \sum_{m=1}^{n-1} \binom{n}{m} 2^{-2m(n-m)} \\
&\leq \frac{1}{2} \sum_{m=1}^{n-1} \binom{n}{m} 2^{-2(n-1)} = 2^{-2n + 1}  \sum_{m=1}^{n-1} \binom{n}{m}
= 2^{-2n+1} (2^n - 2) \\
&< 2^{-2n +1} \to 0 \quad \text{ as } \ n \to \infty.
\end{align*}
Hence $\lim_{n\to\infty} \mbbP_n\big(\mbW_n^0 \cup \mbW_n^n\big) = 1$.
Since $\mbbP_n\big(\mbW_n^0\big) = \mbbP_n\big(\mbW_n^n\big)$
(because $\mu_n\big(\mbW_n^0\big) = 2^{n^2} = \mu_n\big(\mbW_n^n\big)$)
we must have 
$\lim_{n\to\infty} \mbbP_n\big(\mbW_n^0\big)$ = $\lim_{n\to\infty} \mbbP_n\big(\mbW_n^n\big)$ = $1/2$.
This means that for all $\varepsilon > 0$, if $n$ is sufficiently large and $i, j \in [n]$ 
are distinct, then the probability that $R(i)$ holds is in $(1/2 - \varepsilon, 1/2 + \varepsilon)$
and the probability that both $R(i)$ and $R(j)$ hold is also in $(1/2 - \varepsilon, 1/2 + \varepsilon)$.

Let $\mbbG$ be an LBN over $\sigma$.
For $i \in [n]$ let $\mbbP_n^\mbbG(R(i)) := \mbbP_n^\mbbG(\{\mcA \in \mbW_n : \mcA \models R(i)\})$
and $\mbbP_n(R(i)) := \mbbP_n(\{\mcA \in \mbW_n : \mcA \models R(i)\})$.
We use similar notation for $R(i) \wedge R(j)$ in place of $R(i)$.
By Lemma~\ref{independency in LBNs},
if $i, j \in [n]$ are different then 
\[
\mbbP_n^\mbbG(R(i) \wedge R(j)) = 
\mbbP_n^\mbbG(R(i)) \cdot \mbbP_n^\mbbG(R(j)) = \big(\mbbP_n^\mbbG(R(i))\big)^2.
\]
Let $\varepsilon = 0.01$ and suppose that $n$ is large enough that, for all distinct $i, j \in [n]$,
both $\mbbP_n(R(i))$ and $\mbbP_n(R(i) \wedge R(j))$ are in $(1/2 - \varepsilon, 1/2 + \varepsilon)$.
Let $i, j \in [n]$ be distinct.
If $\mbbP_n^\mbbG(R(i)) \in (1/2 - 2\varepsilon, 1/2 + 2\varepsilon)$
then $\mbbP_n^\mbbG(R(i) \wedge R(j)) \in (1/4 - 6\varepsilon, 1/4 + 6\varepsilon)$.
So 
\[
\max\{|\mbbP_n^\mbbG(R(i)) - \mbbP_n(R(i))|, 
|\mbbP_n^\mbbG(R(i) \wedge R(j)) - \mbbP_n(R(i) \wedge R(j))|\} \geq 0.01.
\]
In particular, $(\mbbP_n : n \in \mbbN^+)$
and $(\mbbP_n^\mbbG : n \in \mbbN^+)$ are not asymptotically total variation equivalent.

\section{Proof of Theorem~\ref{characterization of quantifier-free MLNs for colourings}}
\label{proof of characterization of quantifier-free MLNs for colourings}

\noindent
Let $\mbbM$ be a quantifier-free MLN over $\sigma = \{R\}$ where $R$ has arity 1.
For all $n \in \mbbN^+$, let $\mu_n^\mbbM$ and $\mbbP_n^\mbbM$ be as defined
in Definition~\ref{definition of distribution defined by an MLN} and let us use the abbreviations
$\mu_n := \mu_n^\mbbM$ and $\mbbP_n := \mbbP_n^\mbbM$.
We first show that $\mbbM$ can be assumed to have a certain ``normal form''.

\begin{lem}\label{normalized quantifier-free MLN over colourings}
There is a quantifier-free MLN $\mbbM'$ over $\sigma$ such that
\begin{itemize}
\item $\mbbM$ and $\mbbM'$ determine the same probability distribution on $\mbW_n$ for all $n$, and

\item for some $\nu \in \mbbN^+$
\[
\mbbM' = \bigcup_{k=1}^\nu \big\{ (\varphi_{k, 0}(x_1, \ldots, x_k), w_{k, 0}), \ldots, (\varphi_{k, k}(x_1, \ldots, x_k), w_{k, k})\big\}
\]
where, for all $k = 1, \ldots, \nu$ and $s = 0, \ldots, k$, $\varphi_{k, s}(x_1, \ldots, x_k)$ is the formula
\begin{equation}\label{the form of the formulas in M}
\bigwedge_{1 \leq i < j \leq k} x_i \neq x_j \ \wedge \
\bigvee_{\substack{I \subseteq \{1, \ldots, k\} \\ |I| = s}} \bigg( \bigwedge_{i \in I} R(x_i) \ \wedge \ 
\bigwedge_{i \notin I} \neg R(x_i) \bigg).
\end{equation}
\end{itemize}
\end{lem}

\noindent
{\bf Proof.}
By Lemma~\ref{normalized quantifier-free MLN} 
we may assume that for all $(\varphi(\bar{x}), w) \in \mbbM$, $\varphi(\bar{x})$ is a maximal consistent conjunction of
literals and $\varphi(\bar{x})$ implies $x_i \neq x_j$ whenever $x_i$ and $x_j$ are different variables in $\varphi(\bar{x})$.
Since weights are allowed to be 0 we may, without loss of generality, assume that, 
for some $\nu \in \mbbN^+$ and all $k = 1, \ldots, \nu$,
if $\theta(x_1, \ldots, x_k)$ is a maximal consistent conjunction of literals then there is 
$(\theta'(x_1, \ldots, x_k), w) \in \mbbM$ such that $\theta'(x_1, \ldots, x_k)$ is equivalent to $\theta(x_1, \ldots, x_k)$.

It follows that if $(\varphi(\bar{x}), w) \in \mbbM$, where $\bar{x} = (x_1, \ldots, x_k)$ say,
then there is $I \subseteq [k]$ such that $\varphi(\bar{x})$ is equivalent to 
\[
\bigwedge_{1 \leq i < j \leq k} x_i \neq x_j \ \wedge \
\bigwedge_{i \in I} R(x_i) \ \wedge \ 
\bigwedge_{i \notin I} \neg R(x_i).
\]
Suppose that $J \subseteq [k]$ and $|J| = |I|$, and let $\psi(\bar{x})$ be the formula
\[
\bigwedge_{1 \leq i < j \leq k} x_i \neq x_j \ \wedge \
\bigwedge_{i \in J} R(x_i) \ \wedge \ 
\bigwedge_{i \notin J} \neg R(x_i).
\]
So if $\mcA$ is a finite $\sigma$-structure and $\mcA \models \varphi(\bar{a})$, then $\bar{a}$ can be reordered to $\bar{a}'$, say,
so that $\mcA \models \psi(\bar{a}')$, and vice versa.
Hence $|\varphi(\mcA)| = |\psi(\mcA)|$.

Suppose that $|I| = s$. Let $J_1, \ldots J_{\binom{k}{s}}$ enumerate all
subsets of $[k]$ of cardinality $s$.
By assumption we may assume that, for $l = 1, \ldots, \binom{k}{s}$, there is
\[
(\varphi_{k, s, l}(x_1, \ldots, x_k), w_{k, s, l}) \in \mbbM
\]
where $\varphi_{k, s, l}(x_1, \ldots, x_k)$ is the formula
\[
\bigwedge_{1 \leq i < j \leq k} x_i \neq x_j \ \wedge \
\bigwedge_{i \in J_l} R(x_i) \ \wedge \ 
\bigwedge_{i \notin J_l} \neg R(x_i).
\]
By the observation above we have $|\varphi_{k, s, l}(\mcA)| = |\varphi_{k, s, l'}(\mcA)|$ for all $1 \leq l \leq l' \leq \binom{k}{s}$
and every finite $\sigma$-structure $\mcA$.
Therefore we can replace all pairs $(\varphi_{k, s, l}(x_1, \ldots, x_k), w_{k, s, l}) \in \mbbM$
by a single pair $(\varphi_{k, s}(x_1, \ldots, x_k), w_{k, s})$ where 
$\varphi_{k, s}(x_1, \ldots, x_k)$ has the form~(\ref{the form of the formulas in M})
and 
\[
w_{k, s} := \frac{w_{k, s, 1} + \ldots + w_{k, s, \binom{k}{s}}}{\binom{k}{s}}.
\]
The thus obtained MLN will determine the same probability distribution on $\mbW_n$ as $\mbbM$ does.
As we can repeat the procedure for every $k = 1, \ldots, \nu$, this completes the proof.
\hfill $\square$

\bigskip

\noindent
Now we turn to the proof of 
Theorem~\ref{characterization of quantifier-free MLNs for colourings}.
By Lemma~\ref{normalized quantifier-free MLN over colourings} we may assume that 
\[
\mbbM = \bigcup_{k=1}^\nu \big\{ (\varphi_{k, 0}(x_1, \ldots, x_k), w_{k, 0}), \ldots, (\varphi_{k, k}(x_1, \ldots, x_k), w_{k, k})\big\}
\]
where, for all $k = 1, \ldots, \nu$ and $s = 0, \ldots, k$, $\varphi_{k, s}(x_1, \ldots, x_k)$ is the formula
\[
\bigwedge_{1 \leq i < j \leq k} x_i \neq x_j \ \wedge \
\bigvee_{\substack{I \subseteq \{1, \ldots, k\} \\ |I| = s}} \bigg( \bigwedge_{i \in I} R(x_i) \ \wedge \ 
\bigwedge_{i \notin I} \neg R(x_i) \bigg).
\]
Hence $\varphi_{k, s}(x_1, \ldots, x_k)$ expresses that all $x_1, \ldots, x_k$ are different and that exactly $s$ of them are coloured.
We now prove the existence of numbers $\alpha_1, \ldots, \alpha_t \in [0, 1]$ as claimed by 
Theorem~\ref{characterization of quantifier-free MLNs for colourings}.
The proof proceeds by a case analysis.

\subsection*{Case 1}

Suppose that for all $k \in [\nu]$ and all $s, s' \in \{0, \ldots, k\}$ we have $w_{k, s} = w_{k, s'}$.
It follows from the definition of $\mbbP_n$ that for all $\mcA, \mcA' \in \mbW_n$,
$\mbbP_n(\mcA) = \mbbP_n(\mcA')$, so $\mbbP_n$ is the uniform probability distribution.
Then it follows from 
Theorem~\ref{typical proportion of tuples satisfying a formula} 
that there is $\alpha \in (0, 1)$ such that for all $\varepsilon > 0$
\[
\lim_{n\to\infty} \mbbP_n\Big( \Big\{ \mcA \in \mbW_n : 
\frac{|R(\mcA)|}{n} \in (\alpha - \varepsilon, \alpha + \varepsilon) \Big\}\Big) = 1.
\]
(By well-known arguments, using for example 
Lemma~\ref{independent bernoulli trials}, one can see that the above holds with $\alpha = 1/2$.)

\subsection*{Case 2}

Now suppose that for some $k \in [\nu]$ there are $s, s' \in \{0, \ldots, k\}$ such that $w_{k, s} \neq w_{k, s'}$.
Without loss of generality we can assume that there are $s, s' \in \{0, \ldots, \nu\}$ such that $w_{\nu, s} \neq w_{\nu, s'}$,
because otherwise the MLN
\[
\mbbM' = \mbbM \setminus \big\{(\varphi_{\nu, s}(x_1, \ldots, x_\nu), w_{\nu, s})  : s = 0, \ldots, \nu\big\}
\]
would determine the same probability distribution as $\mbbM$ on each $\mbW_n$, so we could consider $\mbbM'$ instead
(and redefine $\nu := \nu - 1$).

\subsection*{Case 2a}

Suppose that $\nu \geq 2$.
For $m, k \in \mbbN^+$ let 
\[
(m)_k := m(m-1) \cdot \ldots \cdot (m-k+1), \ (0)_k := 0, \ (m)_0 := 1, \ (0)_0 := 1.
\]
For each integer $l \geq 2$ there is a polynomial $p_l$ of degree $l-1$ such that for all $n \in \mbbN^+$,
$p_l(n)$ is the number of ordered $l$-tuples of elements from $[n]$ such that at least two different entries in the $l$-tuple are equal. 

Suppose that $M \subseteq [n]$ and $|M| = m$. 
The number of ordered $k$-tuples of distinct elements from $[n]$ such that exactly $s$ of them belong to $M$ is
\[
\binom{k}{s} (m)_s (n-m)_{k-s}
\]
because we can choose the places in the $k$-tuple which are to hold the elements from $M$ in $\binom{k}{s}$ ways.
We have $(m)_s = m^s - p_s(m)$ and $(n-m)_{k-s} = (n-m)^{k-s} - p_{k-s}(n-m)$, so we get

\begin{align}\label{the number of s-coloured k-tuples of distinct elements}
&\binom{k}{s} (m)_s (n-m)_{k-s} = \\
&\binom{k}{s}\Big(m^s - p_s(m)\Big) \Big((n-m)^{k-s} - p_{k-s}(n-m)\Big) = \nonumber \\
&\binom{k}{s}m^s(n-m)^{k-s} \ - \nonumber \\ 
&\binom{k}{s}\Big[ m^s p_{k-s}(n-m) + p_s(m)(n-m)^{k-s} - p_s(m)p_{k-s}(n-m) \Big] =  \nonumber \\
&\binom{k}{s}m^s(n-m)^{k-s} - O\big(n^{k-1}\big). \nonumber
\end{align}

\noindent
By~(\ref{the number of s-coloured k-tuples of distinct elements}) we have
\begin{align}\label{s-coloured k-tuples divided by n-k}
&\frac{\binom{k}{s} (m)_s (n-m)_{k-s}}{n^k} = 
\frac{\binom{k}{s} m^s (n-m)^{k-s} - O\big(n^{k-1}\big)}{n^k} =  \\
&\frac{\binom{k}{s} (m)^s (n-m)^{k-s}}{n^s n^{k-s}} - O\Big(\frac{1}{n}\Big) = \nonumber \\
&\binom{k}{s}\Big(\frac{m}{n}\Big)^s \Big(1 - \frac{m}{n}\Big)^{k-s} - O\Big(\frac{1}{n}\Big). \nonumber
\end{align}
In both (\ref{the number of s-coloured k-tuples of distinct elements}) and 
(\ref{s-coloured k-tuples divided by n-k}) 
the error bounds are uniform in $m$ for $0 \leq m \leq n$.
For all $\alpha \in \mbbR$, define
\[
f_k(\alpha) := \sum_{s=0}^k w_{k, s} \binom{k}{s}\alpha^s(1 - \alpha)^{k-s}
\]
so $f_k$ is a polynomial of degree $k$ which takes only nonnegative values when restricted to $[0, 1]$.
By assumption there are $s, s' \in \{0, \ldots, \nu\}$ such that $w_{\nu, s} \neq w_{\nu, s'}$.
It follows from Proposition~\ref{when the weighted binomial distribution is constant}
that $f_\nu$ is not constant, and as $f_\nu$ is a polynomial it is still not constant when restricted to $[0, 1]$.

Recall that $M \subseteq [n]$ and $|M| = m$. 
Suppose that $\mcA \in \mbW_n$ and $R^\mcA = M$.
By~(\ref{the number of s-coloured k-tuples of distinct elements})
and~(\ref{s-coloured k-tuples divided by n-k})
we get
\begin{align*}
&\mu_n(\mcA) = 
\exp_2\bigg(\sum_{k=1}^\nu \sum_{s=0}^k w_{k, s}\binom{k}{s} (m)_s (n-m)_{k-s}\bigg) = \\
&\exp_2\bigg(\sum_{k=1}^\nu n^k \sum_{s=0}^k w_{k, s} \frac{\binom{k}{s} (m)_s (n-m)_{k-s}}{n^k}\bigg) = \\
&\exp_2\bigg(\sum_{k=1}^\nu n^k \sum_{s=0}^k 
\bigg[w_{k, s}\binom{k}{s}\Big(\frac{m}{n}\Big)^s \Big(1 - \frac{m}{n}\Big)^{k-s} - O\Big(\frac{1}{n}\Big) \bigg] \bigg) = \\
&\exp_2\bigg(\sum_{k=1}^\nu n^k \bigg[ \sum_{s=0}^k 
w_{k, s}\binom{k}{s}\Big(\frac{m}{n}\Big)^s \Big(1 - \frac{m}{n}\Big)^{k-s} - O\Big(\frac{1}{n}\Big) \bigg]\bigg) = \\
&\exp_2\bigg(\sum_{k=1}^\nu n^k \Big(f_k\Big(\frac{m}{n}\Big) - O\Big(\frac{1}{n}\Big) \Big) \bigg).
\end{align*}

For all $m = 0, 1, \ldots, n$ define
\[
\mbW_n^m := \big\{ \mcA \in \mbW_n : |R^\mcA| = m \big\}.
\]
Since we can choose $M \subseteq [n]$ such that $|M| = m$ in $\binom{n}{m}$ ways it follows that
\begin{align}\label{characterization of probability of W-n-m generally}
\mu_n\big(\mbW_n^m\big) &= 
\binom{n}{m} \exp_2\bigg(\sum_{k=1}^\nu n^k \Big(f_k\Big(\frac{m}{n}\Big) - O\Big(\frac{1}{n}\Big) \Big) \bigg) \\
&= \binom{n}{m} \exp_2\bigg(
n^\nu \Big(f_\nu\Big(\frac{m}{n}\Big) - O\Big(\frac{1}{n}\Big) \Big) + O\big(n^{\nu - 1}\big) \bigg). \nonumber
\end{align}

Let $\beta$ be the maximal value that $f_\nu$ takes on $[0, 1]$, that is, $\beta = \sup\{f_\nu(\alpha) : \alpha \in [0, 1]\}$.
Since $f_\nu$ takes only nonnegative values on $[0, 1]$ and is not constant it follows that $\beta > 0$.
Let $\alpha_1, \ldots, \alpha_t \in [0, 1]$ enumerate all $\alpha \in [0, 1]$ such that $f_\nu(\alpha) = \beta$.
Let $\varepsilon > 0$.

As $f_\nu$ is continuous we can choose $\delta > 0$ small enough such that if
$\alpha \in [0, 1]$ and $\alpha \notin \bigcup_{i=1}^t (\alpha_i - \varepsilon, \alpha_i + \varepsilon)$
then $f_\nu(\alpha) < \beta - \delta$.
Suppose that $\nu \geq 2$.

\begin{align*}
& \mbbP_n\Big( \Big\{ \mcA \in \mbW_n : 
\frac{|R(\mcA)|}{n} \notin \bigcup_{i=1}^t (\alpha_i - \varepsilon, \alpha_i + \varepsilon) \Big\}\Big) = \\
&\frac{\mu_n\Big( \Big\{ \mcA \in \mbW_n : 
\frac{|R(\mcA)|}{n} \notin \bigcup_{i=1}^t (\alpha_i - \varepsilon, \alpha_i + \varepsilon) \Big\}\Big)}
{\mu_n\big(\mbW_n\big)} \leq \\
&\frac{\mu_n\Big( \Big\{ \mcA \in \mbW_n : 
\frac{|R(\mcA)|}{n} \notin \bigcup_{i=1}^t (\alpha_i - \varepsilon, \alpha_i + \varepsilon) \Big\}\Big)}
{\mu_n\big(\mbW_n^{\lfloor \alpha_1 n \rfloor}\big)} = \\
&\frac{\sum \big\{ \mu_n\big(\mbW_n^m\big)  : 
m \in [n] \text{ and } \frac{m}{n} \notin \bigcup_{i=1}^t (\alpha_i - \varepsilon, \alpha_i + \varepsilon)\big\}}
{\mu_n\big(\mbW_n^{\lfloor \alpha_1 n \rfloor}\big)} = \\
&\frac{\sum \big\{ 
\binom{n}{m} \exp_2\big( n^\nu \big(f_\nu\big(\frac{m}{n}\big) - o(1) \big) + O\big(n^{\nu - 1}\big) \big) : 
m \in [n] \text{ and } \frac{m}{n} \notin \bigcup_{i=1}^t (\alpha_i - \varepsilon, \alpha_i + \varepsilon)\big\}}
{\binom{n}{\lfloor \alpha_1 n\rfloor} 
\exp_2\Big( n^\nu \big(f_\nu\big(\frac{\lfloor \alpha_1 n \rfloor}{n}\big) - o(1) \big) + O\big(n^{\nu - 1}\big) \Big) } \leq \\
&\frac{e^{cn} \exp_2\big( n^\nu \big(\beta - \delta - o(1) \big) + O\big(n^{\nu - 1}\big)\big) }
{\binom{n}{\lfloor \alpha_1 n\rfloor} 
\exp_2\Big( n^\nu \big(f_\nu\big(\frac{\lfloor \alpha_1 n \rfloor}{n}\big) - o(1) \big) + O\big(n^{\nu - 1}\big) \Big)} =
\qquad \qquad \text{ for some $c > 0$} \\
&\frac{e^{cn}}
{\binom{n}{\lfloor \alpha_1 n\rfloor} 
\exp_2\Big( n^\nu \big(f_\nu\big(\frac{\lfloor \alpha_1 n \rfloor}{n}\big) - \beta + \delta \pm o(1) \big) \pm O\big(n^{\nu - 1}\big) \Big)}
\end{align*}

\noindent
Since $\nu \geq 2$ and $f_\nu\big(\frac{\lfloor \alpha_1 n \rfloor}{n}\big) \to \beta$ as $n \to \infty$ it follows that the last expression above tends to 0
as $n \to \infty$. Hence 
\[
\lim_{n\to\infty} \mbbP_n\Big( \Big\{ \mcA \in \mbW_n : 
\frac{|R(\mcA)|}{n} \notin \bigcup_{i=1}^t (\alpha_i - \varepsilon, \alpha_i + \varepsilon) \Big\}\Big) = 0
\]
and we are done in the case when $\nu \geq 2$.

\subsection{Case 2b}

Suppose that $\nu = 1$.
Let us rename $w_1 := w_{1, 1}$ and $w_0 := w_{1, 0}$.
Let 
\[
p := \frac{2^{w_1}}{2^{w_1} + 2^{w_0}} \quad \text{ so } \quad 
1 - p = \frac{2^{w_0}}{2^{w_1} + 2^{w_0}}.
\]
Observe that, by the binomial theorem,
\[
\big(2^{w_1} + 2^{w_0}\big)^n = \sum_{m=0}^n \binom{n}{m} 2^{mw_1} \cdot 2^{(n-m)w_0}
= \sum_{m=0}^n \binom{n}{m} 2^{w_1m + w_0(n-m)}.
\]
For all integers $0 \leq m \leq n$ and $\mcA \in \mbW_n^m$ 
we have $\mu_n(\mcA) = 2^{w_1 m + w_2 (n-m)}$ and hence
\begin{align}\label{P is a Bernoulli distribution}
&\mbbP_n(\mcA) = \frac{2^{w_1 m + w_2 (n-m)}}{\sum_{k=0}^n \binom{n}{k} 2^{w_1 k + w_2 (n-k)}}
= \frac{2^{w_1 m + w_2 (n-m)}}{\big(2^{w_1} + 2^{w_0}\big)^n} \\
&\frac{2^{w_1 m}}{\big(2^{w_1} + 2^{w_0}\big)^m} \cdot 
\frac{2^{w_0 (n-m)}}{\big(2^{w_1} + 2^{w_0}\big)^{n-m}}
= p^m (1-p)^{n-m}. \nonumber
\end{align}
Let $Z : \mbW_n \to \mbbN$ be defined by $Z(\mcA) = |R(\mcA)|$, so if $\mcA \in \mbW_n^m$ then $Z(\mcA) = m$.
Then~(\ref{P is a Bernoulli distribution}) implies that $Z$ corresponds to a sum of $n$ independent 0/1-valued 
random variables with probability $p$ of having value 1.
It follows from 
Lemma~\ref{independent bernoulli trials}
that for every $\varepsilon > 0$,
\[
\lim_{n\to\infty} \mbbP_n\Big( \Big\{ \mcA \in \mbW_n : 
\frac{|R(\mcA)|}{n} \notin (p - \varepsilon, p + \varepsilon) \Big\}\Big) 
= 0.
\]
This completes the proof in the case 2b and hence the proof of
Theorem~\ref{characterization of quantifier-free MLNs for colourings}.

\section{Proof of Theorem~\ref{limits of MLNs}}\label{proof of limits of MLNs}

\noindent
Suppose that $\sigma = \{R\}$ where $R$ has arity 1.
Let $\alpha \in (0, \infty)$, $p_n = n^{-\alpha}$ for all $n \in \mbbN^+$,
and let $\mbbG$ be as described in Theorem~\ref{limits of MLNs}.
Then, for all $n \in \mbbN^+$
and $\mcA \in \mbW_n$, 
\[
\mbbP_n^\mbbG(\mcA) = p_n^{|R(\mcA)|} (1 - p_n)^{n - |R(\mcA)|}.
\]
Fix an arbitrary quantifier-free MLN
\[
\mbbM = \big\{(\varphi_1(\bar{x}_1), w_1), \ldots, (\varphi_t(\bar{x}_t), w_t)\big\}.
\]
For all $n \in \mbbN^+$ let $\mcE_n, \mcF_n \in \mbW_n$ be structures such that
$R(\mcE_n) = \es$ and $R(\mcF_n) = [n]$, so $\mcE_n \models \forall x \neg R(x)$ 
and $\mcF_n \models \forall x R(x)$.
Then
\begin{equation}\label{F-n over E-n}
\frac{\mbbP_n^\mbbM(\mcF_n)}{\mbbP_n^\mbbM(\mcE_n)} = 
\frac{\mu_n^\mbbM(\mcF_n)}{\mu_n^\mbbM(\mcE_n)} = 
\exp_2\Big(\sum_{i=1}^t w_i\big(|\varphi_i(\mcF_n)| - |\varphi_i(\mcE_n)|\big)\Big).
\end{equation}
Let $\varphi(x_1, \ldots, x_k)$ be any quantifier-free formula and let $a_1, \ldots, a_k \in [n]$.
For the rest of the argument suppose that $n > \max\{1, k\}$.
Then $\mcE_n \models \neg R(a_i)$ and $\mcF_n \models R(a_i)$ for all $i = 1, \ldots, k$.
Hence the truth value of $\varphi(a_1, \ldots, a_k)$ in $\mcE_n$, respectively $\mcF_n$,
depends only on whether $a_i = a_j$ or not, for $i, j = 1, \ldots, k$.

Let $\pi$ be a partition of $[k]$.
Let us say that $(a_1, \ldots, a_k) \in [n]^k$ {\em realizes} $\pi$ if for all $i, j = 1, \ldots, k$,
$a_i = a_j$ if and only if $i$ and $j$ belong to the same part of $\pi$.
If $\pi$ has exactly $m$ parts then the number of $k$-tuples from $[n]^k$ that realize $\pi$
is 
\[
(n)_m := n(n-1) \cdot \ldots \cdot (n-m+1).
\]
Let $\Pi_k$ be the (finite) set of partitions of $[k]$.
For each $\pi \in \Pi_k$, define
\[
\xi_\pi^e = 
\begin{cases}
0 \ \text{ if  $\mcE_n \models \neg\varphi(a_1, \ldots, a_k)$ whenever $(a_1, \ldots, a_k) \in [n]^k$ realizes $\pi$,} \\
1 \ \text{ if $\mcE_n \models \varphi(a_1, \ldots, a_k)$ whenever $(a_1, \ldots, a_k) \in [n]^k$ realizes $\pi$.}
\end{cases}
\]
and 
\[
\xi_\pi^f = 
\begin{cases}
0 \ \text{ if  $\mcF_n \models \neg\varphi(a_1, \ldots, a_k)$ whenever $(a_1, \ldots, a_k) \in [n]^k$ realizes $\pi$,} \\
1 \ \text{ if $\mcF_n \models \varphi(a_1, \ldots, a_k)$ whenever $(a_1, \ldots, a_k) \in [n]^k$ realizes $\pi$.}
\end{cases}
\]
It follows that if $m_\pi$ denotes the number of parts of $\pi \in \Pi_k$, then
\begin{equation}\label{form of varphi(E-n)}
|\varphi(\mcE_n)| = \sum_{\pi \in \Pi_k} \xi_\pi^e \cdot (n)_{m_\pi} \quad \text{ and } \quad
|\varphi(\mcF_n)| = \sum_{\pi \in \Pi_k} \xi_\pi^f \cdot (n)_{m_\pi}.
\end{equation}
Consequently, $|\varphi(\mcE_n)| - |\varphi(\mcF_n)|$ is a polynomial in $n$.
Now suppose that
\[
n > \max\{1, |\bar{x}_1|, \ldots, |\bar{x}_t|\}.
\]
Then the above conclusion can be applied to each one of $\varphi_1(\bar{x}_1), \ldots, \varphi_t(\bar{x}_t)$ and
it follows that 
\[
Q(n) := \sum_{i=1}^t w_i\big(|\varphi_i(\mcF_n)| - |\varphi_i(\mcE_n)|\big)
\]
is a polynomial in $n$. Together with~(\ref{F-n over E-n}) we get
\begin{equation}\label{log of the fraction for M}
\log_2 \bigg( \frac{\mbbP_n^\mbbM(\mcF_n)}{\mbbP_n^\mbbM(\mcE_n)} \bigg) = 
Q(n).
\end{equation}

Since $\mbbP_n^\mbbG(\mcF_n) = p_n^n$ and $\mbbP_n^\mbbG(\mcE_n) = (1 - p_n)^n$
we get 
\[
\frac{\mbbP_n^\mbbG(\mcF_n)}{\mbbP_n^\mbbG(\mcE_n)} = \bigg(\frac{p_n}{1 - p_n}\bigg)^n = 
\bigg(\frac{n^{-\alpha}}{1 - n^{-\alpha}}\bigg)^n = \frac{1}{(n^\alpha - 1)^n}
= (n^\alpha - 1)^{-n}.
\]
Hence 
\begin{equation}\label{log of the fraction of G}
\log_2\bigg(\frac{\mbbP_n^\mbbG(\mcF_n)}{\mbbP_n^\mbbG(\mcE_n)}\bigg) = 
- n \log_2 \big(n^\alpha - 1\big).
\end{equation}

Towards a contradiction, suppose that there is an infinite strictly increasing sequence $n_j \in \mbbN^+$,
$j \in \mbbN$, such that
$\mbbP_{n_j}^\mbbM = \mbbP_{n_j}^\mbbG$ for all $j \in \mbbN$.
From~(\ref{log of the fraction for M}) and~(\ref{log of the fraction of G}) we get
\begin{equation}\label{new expression for Q}
Q(n_j) = - n_j \log_2(n_j^\alpha - 1)
\end{equation}
if $j$ is large enough, so (using that $0 < \alpha$ in the final equality)  we get
\begin{equation}\label{limit of Q over n-2}
\lim_{j \to \infty} \frac{Q(n_j)}{n_j^2} = \lim_{j \to \infty} \frac{\log_2\big(n_j^\alpha - 1)}{n_j} = 0.
\end{equation}
Hence $\lim_{x \to \infty} \frac{Q(x)}{x^2} = 0$ and therefore the degree of $Q(x)$ can be at most $1$.
Thus $Q(x) = ax + b$ for some constants $a$ and $b$, and hence
\[
\lim_{j \to \infty} \frac{Q(n_j)}{n_j} = a.
\]
Combining this with~(\ref{new expression for Q}) gives
\[
\lim_{j \to \infty} \big( - \log_2(n_j^\alpha - 1) \big) = 
\lim_{j \to \infty} \frac{- n_j \log_2(n_j^\alpha - 1)}{n_j} = a.
\]
But as $\alpha > 0$ and $\lim_{j \to \infty} n_j = \infty$ we have 
$\lim_{j \to \infty} \big( - \log_2(n_j^\alpha - 1) \big) = -\infty$, contradicting the above.
This concludes the proof of 
Theorem~\ref{limits of MLNs}.

\section{Proof of Theorem~\ref{probability of max degree Delta tends to 0}}
\label{proof of probability of max degree Delta tends to 0}

\noindent
Let $\varphi(x_1, \ldots, x_{\Delta + 2})$ be the formula
\[
\bigvee_{2 \leq i < j \leq \Delta + 2} x_i = x_j \ \vee \ \bigvee_{i = 2}^{\Delta + 2} \neg R(x_1, x_i),
\]
and let 
\[
\mbbM := \big\{(\varphi(x_1, \ldots, x_{\Delta + 2}), w)\big\} \quad \text{ where } w \geq 0. 
\]
For all $n \in \mbbN^+$ and $m \in \mbbN$ let
\[
\mbOm_n^{m} := \big\{ \mcA \in \mbW_n : \text{ $\mcA$ has maximum degree at most $m$} \big\}.
\]
Also let $\mu_n := \mu_n^\mbbM$ and $\mbbP_n := \mbbP_n^\mbbM$.
If $\mcA \in \mbW_n$ and $a \in [n]$ then let 
\[
N_\mcA(a) := \{a\} \cup \{b \in [n] : \text{ $b$ is adjacent to $a$}\}.
\]
Fix any $\Delta \in \mbbN^+$.
If $n \geq \Delta + 3$, $i \in [n]$, and $\mcA \in \mbOm_n^{\Delta}$, then define
$F_i(\mcA) \in \mbW_n$ to be the graph that is obtained by
adding edges between $i$ and the first $\Delta + 2$ vertices (in $[n]$) that are different from $i$.
Then
\begin{enumerate}
\item in the graph $F_i(\mcA)$, the vertex $i$ has degree $\geq \Delta + 2$ and it is the only vertex with degree $\geq \Delta + 2$, 
\item in the graph $F_i(\mcA)$, the vertex $i$ has degree $\leq 2\Delta + 2$, 
\item in the graph $F_i(\mcA)$, at most $\Delta + 3$ vertices have degree $\geq \Delta + 1$, and
\item in the graph $F_i(\mcA)$, if $v$ is a vertex with degree $\geq \Delta + 1$ and $v \neq i$, then its degree is exactly $\Delta + 1$.
\end{enumerate}
Let $F_i\big(\mbOm_n^{\Delta}\big) := \big\{ F_i(\mcA) : \mcA \in \mbOm_n^{\Delta}\big\}$.

\begin{lem}\label{F is almost injective}
For all sufficiently large $n$ and all $i \in [n]$,
$\big|F_i\big(\mbOm_n^{\Delta}\big)\big| \geq \big|\mbOm_n^{\Delta}\big| \Big/ 2^{^{\binom{2\Delta + 3}{2}}}$.
\end{lem}

\noindent
{\bf Proof.}
Let $\mcA, \mcA' \in \mbOm_n^{\Delta}$, $i \in [n]$, and suppose that $F_i(\mcA) = F_i(\mcA')$.
Then $i$ has exactly the same neighbours in $F_i(\mcA)$ as in $F_i(\mcA')$, so
$N_{F_i(\mcA)}(i) = N_{F_i(\mcA')}(i)$.
For every vertex $v$ outside of $N_{F_i(\mcA)}(i) = N_{F_i(\mcA')}(i)$,
$v$ has the same neighbours (by definition of $F_i$) in $\mcA$ as in $\mcA'$.
Hence $\mcA$ and $\mcA'$ can only differ on $N_{F_i(\mcA)}(i) = N_{F_i(\mcA')}(i)$
where $|N_{F_i(\mcA)}(i)| \leq 2\Delta + 3$.
Since there are at most $s := 2^{^{\binom{2\Delta + 3}{2}}}$ (undirected) graphs (without loops) on a vertex set with 
at most $2\Delta + 3$ vertices there can be at most $s$ distinct members of $\mbOm_n^{\Delta}$ that are
mapped by $F_i$ to the same graph.
Hence $\big|F_i\big(\mbOm_n^{\Delta}\big)\big| \geq \big|\mbOm_n^{\Delta}\big| \big/ s$.
\hfill $\square$

\begin{lem}\label{F is almost injective, second part}
For all sufficiently large $n$, all $i, j \in [n]$, and all $\mcA, \mcA' \in \mbOm_n{\Delta}$,
if $i \neq j$ then $F_i(\mcA) \neq F_j(\mcA')$.
Hence $F_i\big(\mbOm_n^{\Delta}\big) \cap F_j\big(\mbOm_n^{\Delta}\big) = \es$ if $i \neq j$.
\end{lem}

\noindent
{\bf Proof.}
By (1) above, $i$ is the unique vertex in $F_i(\mcA)$ that has more than $\Delta + 1$ neighbours,
and $j$ is the unique vertex in $F_j(\mcA')$ that has more than $\Delta + 1$ neighbours.
So if $i \neq j$ then $F_i(\mcA) \neq F_j(\mcA')$.
\hfill $\square$

\begin{lem}\label{number of tuples violating the constraint}
Suppose that $n \geq \Delta + 3$, $i \in [n]$, and $\mcA \in \mbOm_n^{\Delta}$. 
The number of ordered tuples $(a_1, \ldots, a_{\Delta + 2}) \in [n]^{\Delta + 2}$ such that 
$F_i(\mcA) \models \neg \varphi(a_1, \ldots, a_{\Delta + 2})$ is at most
$\tau := \binom{2\Delta + 2}{\Delta + 1}! + (\Delta + 2)(\Delta + 1)!$.
Hence $\mu_n\big(F_i(\mcA)\big) \geq 2^{w(n^{\Delta + 2} - \tau)}$.
\end{lem}

\noindent
{\bf Proof.}
We have $\mcA \models \neg \varphi(a_1, \ldots, a_{\Delta + 2})$ if and only if 
all $a_2, \ldots, a_{\Delta + 2}$ are different and $a_1$ is adjacent to all $a_2, \ldots, a_{\Delta + 2}$,
so $a_1$ must be a vertex with degree at least $\Delta + 1$.
Suppose that $\mcA \in \mbOm_n^{\Delta}$ and $F_i(\mcA) \models \neg \varphi(a_1, \ldots, a_{\Delta + 2})$.
Then, by~(3) above, we have at most $\Delta + 3$ choices of $a_1$, one of which is $a_1 = i$.
If $a_1 = i$ then (by~(2)) $a_1$ has at most $2\Delta + 2$ neighbours (in $F_i(\mcA)$) and we can choose $\Delta + 1$ of them 
as $a_2, \ldots, a_{\Delta + 2}$ in at most $t := \binom{2\Delta + 2}{\Delta + 1}$ ways and order them in at most $t!$ ways. 
If $a_1 \neq i$ then (by~(4)) then $a_1$ has exactly $\Delta + 1$ neighbours (in $F_i(\mcA)$)
which can be ordered in $(\Delta + 1)!$ ways. 
Hence the number of $(a_1, \ldots, a_{\Delta + 2}) \in [n]^{\Delta + 2}$ such that
$F_i(\mcA) \models \neg \varphi(a_1, \ldots, a_{\Delta + 2})$
is at most $t! + (\Delta + 2)(\Delta + 1)!$.
\hfill $\square$

\medskip

\noindent
If $\mcA \in \mbOm_n^{\Delta}$ then no $(\Delta + 2)$-tuple of vertices from $[n]$ falsifies 
$\varphi(x_1, \ldots, x_{\Delta + 2})$ in $\mcA$, so
\begin{equation}\label{mu of Omega}
\mu_n\big(\mbOm_n^{\Delta}\big) = \big|\mbOm_n^{\Delta}\big| \cdot 2^{wn^{\Delta + 2}}.
\end{equation}
By Lemmas~\ref{F is almost injective} and~\ref{F is almost injective, second part} we get
\begin{equation}\label{cardinality of F of the Omegas}
\bigg| \bigcup_{i=1}^n F_i\big(\mbOm_n^{\Delta}\big) \bigg| \ \geq \ 
\sum_{i=1}^n \big|F_i\big(\mbOm_n^{\Delta}\big)\big| \geq  
\sum_{i=1}^n \frac{\big|\mbOm_n^{\Delta}\big|}{2^{^{\binom{2\Delta + 3}{2}}}} = 
\frac{n \big|\mbOm_n^{\Delta}\big|}{2^{^{\binom{2\Delta + 3}{2}}}}.
\end{equation}
By~(\ref{cardinality of F of the Omegas}) and Lemma~\ref{number of tuples violating the constraint}
we get the following, where 
$\tau = \binom{2\Delta + 2}{\Delta + 1}! + (\Delta + 2)(\Delta + 1)!$:
\begin{align}\label{mu of F of the Omegas}
\mu_n\bigg(\bigcup_{i=1}^n F_i\big(\mbOm_n^{\Delta}\big)\bigg) 
\geq \frac{n \big|\mbOm_n^{\Delta}\big|}{2^{^{\binom{2\Delta + 3}{2}}}} \cdot 2^{w(n^{\Delta + 2} - \tau)}.
\end{align}
Finally, by combining~(\ref{mu of Omega}) and~(\ref{mu of F of the Omegas}),
\begin{align*}
\mbbP_n\big(\mbOm_n^{\Delta}\big) &= \frac{\mu_n\big(\mbOm_n^{\Delta}\big)}{\mu_n\big(\mbW_n\big)} 
\leq \frac{\mu_n\big(\mbOm_n^{\Delta}\big)}{\mu_n\bigg(\bigcup_{i=1}^n F_i\big(\mbOm_n^{\Delta}\big)\bigg)} 
\leq \frac{\big|\mbOm_n^{\Delta}\big| \cdot 2^{wn^{\Delta + 2}}}
{\frac{n \big|\mbOm_n^{\Delta}\big|}{2^{^{\binom{2\Delta + 3}{2}}}} \cdot 2^{w(n^{\Delta + 2} - \tau)}} \\
&= \frac{2^{^{ w\tau + \binom{2\Delta + 3}{2}}}}{n} \ \to \ 0 \quad \text{ as } \ n \to \infty.
\end{align*}
This completes the proof of
Theorem~\ref{probability of max degree Delta tends to 0}
and note that we get the above conclusion independently of what the fixed $w$ and $\Delta$ are.

\end{document}